\documentclass[10pt]{article}

\usepackage[margin=1in]{geometry}
\usepackage[T1]{fontenc}
\usepackage[utf8]{inputenc}
\usepackage{lmodern}           
\usepackage{microtype}
\usepackage{amsmath,amssymb,amsfonts,amsthm}
\usepackage{mathtools}
\usepackage{graphicx}
\usepackage{subcaption}        
\usepackage{booktabs}
\usepackage{siunitx}
\usepackage{xcolor}
\usepackage{authblk}           

\theoremstyle{plain}
\newtheorem{theorem}{Theorem}

\theoremstyle{definition}
\newtheorem{definition}{Definition}
\theoremstyle{remark}

\newtheorem{example}{Example}

\usepackage[ruled]{algorithm2e}
\usepackage{amsmath}
\usepackage{amssymb}
\usepackage{comment}
\usepackage{color}
\usepackage{graphicx}
\usepackage{accents}
\usepackage[colorlinks=true]{hyperref}
\usepackage{scalefnt}
\usepackage{tikz}
\usepackage{url}

\newcommand{\dst}{\displaystyle}
\newcommand{\cK}{{\cal K}}
\newcommand{\ubar}[1]{\underaccent{\bar}{#1}}

\usepackage[charsperline=0]{jlcode}
\addlitjlmacros{@constraint}{@constraint}{11}
\addlitjlmacros{@variable}{@variable}{9}
\addlitjlmacros{@objective}{@objective}{10}

\DeclareMathOperator*{\E}{\mathbb{E}}
\DeclareMathOperator*{\argmin}{\text{arg} \min}

\usetikzlibrary{
    decorations.pathmorphing, 
    arrows, 
    shapes.geometric,
}
\tikzset{
    ->,
    >=stealth',
    shorten >= 1pt,
    auto,
    node distance = 1cm, thick,
    bellman/.style={circle, draw},
    square/.style={regular polygon sides = 4, draw},
    textnode/.style={}
}
\newcommand{\drawsquiggle}[2]{
    \draw[
        -stealth,
        decoration = {
            snake,
            amplitude = 0.4mm,
            segment length = 2mm,
            post length = 1mm},
        decorate
    ] (#1) -- (#2);
}

\usetikzlibrary{positioning,arrows.meta,decorations.pathmorphing,calc}

\newcommand{\revision}[1]{\textcolor{black}{#1}}

\title{MDP modeling for multi-stage stochastic programs}
\author[1]{David P.\ Morton}
\author[2]{Oscar Dowson}
\author[3]{Bernardo K.\ Pagnoncelli}
\affil[1]{{\small Department of Industrial Engineering and Management Sciences, Northwestern University, Evanston, IL, USA; \url{david.morton@northwestern.edu}}}
\affil[2]{{\small Dowson Farms, Auckland, New Zealand; \url{oscar@dowsonfarms.co.nz}}}
\affil[3]{{\small SKEMA Business School, Universit{\'e} C{\^o}te d’Azur, Lille, France; \url{bernardo.pagnoncelli@skema.edu}}}

\date{April 2026} 

\begin{document}
\maketitle

\begin{abstract}
\noindent 
We study a class of multi-stage stochastic programs, which incorporate modeling features from Markov decision processes (MDPs). This class includes structured MDPs with continuous state and action spaces. We extend policy graphs to include decision-dependent uncertainty for one-step transition probabilities as well as a limited form of statistical learning. We focus on the expressiveness of our modeling approach, illustrating ideas with a series of examples of increasing complexity. As a solution method, we develop new variants of stochastic dual dynamic programming, including approximations to handle non-convexities. 
\end{abstract}


\section{Introduction}\label{sec:introduction}


\revision{We study sequential decision problems under uncertainty, in which an agent takes an action at each time period that influences both the present (e.g., a reward is received) and the future (e.g., the {state will change} in the next time period). Multiple modeling approaches handle sequential decision-making under uncertainty; see, e.g., \cite{powell2014clearing}. The two most relevant to our work are multi-stage stochastic programming (MSP) \cite{birge2011introduction,king_wallace_2012,shapiro2021lectures} and Markov decision processes (MDPs) \cite{bellman_dynamic_1957,de_farias_linear_2003,howard_dynamic_1960,powell_approximate_2011,puterman2014markov}. From the perspective of MSP, we extend stochastic programs to incorporate attractive modeling features of MDPs, including a type of {\it decision-dependent uncertainty} and a form of {\it statistical learning}. These extensions {can be solved} by variants of stochastic dual dynamic programming~(SDDP), a workhorse algorithm for large-scale MSP with many stages; see the seminal work of~\cite{pereira_multi-stage_1991}, as well as the surveys of \cite{fullner2021stochastic} and \cite[Ch.~2]{seranillaThesis2023}. From the perspective of MDPs, we provide algorithms for solving a class of offline problems---which we call CACS MDPs---with continuous action and continuous state spaces, avoiding the need for discretization.}


\revision{A discrete time, action, and state MDP has states~$\mathcal{S}$, state-dependent actions $\mathcal{A}_s$, and a transition function, $\mathbb{P}(s^\prime | s, a)$, which gives the one-step probability of moving to state $s^\prime$ conditioned on taking action $a$ in state~$s$. As the notation indicates, $\mathbb{P}(s^\prime | s, a)$ depends on action $a$. The solution to an MDP is a policy, $\pi(a|s)$, that gives the (typically degenerate) probability of taking action~$a$ in state $s$, and that maximizes a function of the rewards 
from a starting state, which is governed by distribution~$\mu(s)$.}

\revision{Many authors have relaxed the classical assumption that an MDP's parameters are known.  Wiesemann et al.\ \cite{wiesemann2013robust} deal with uncertainty in one-step transition probabilities using robust optimization. Wang et al.\ \cite{wang2021learning} use covariates instead of process trajectories to deal with uncertain transition and reward functions. Reinforcement learning (RL) deals {with sequential decision problems}  \cite{barto2017some,li2019reinforcement} by simultaneously estimating (i.e., learning) an optimal policy and the model's transition and reward functions.}


\revision{When the {action and state} spaces are finite and of modest size, exact dynamic programming algorithms can solve MDPs. Backward induction is used in finite-horizon problems, and in infinite-horizon problems policy iteration, value iteration, and their variants are used~\cite{puterman2014markov}. As the size of the spaces grows, the cost-to-go functions are approximated, either using sampling~\cite{chang2013simulation}, assuming a specific functional form (e.g., separable approximations~\cite{topaloglu2006dynamic}), using basis functions (e.g., \cite{maxwell2010approximate}), or via deep learning~\cite{le2018deep}. While there are exceptions (e.g., \cite{lan2023policy}), most work in computational MDPs uses discrete actions and states.}


\revision{We extend MSP with MDP modeling constructs, incorporating decision-dependent one-step transition probabilities and learning from a set of hypothesized models. Doing so frames stochastic programming as a \emph{modeling} and \emph{solution technique} for CACS MDPs. 
Using \textit{policy graphs}~\cite{dowson_policy_2018}, i.e., stochastic programs defined on Markov chains, we show that a broad class of MDPs can be modeled via MSP and approximately solved using SDDP. An SDDP algorithm exploits convexity and duality to construct a polyhedral outer approximation of {the} cost-to-go functions. When states and actions are continuous---and in some restricted cases, discrete---and the transition and cost functions are convex, that approximation is asymptotically optimal \cite{philpott_convergence_2008,girardeau_convergence_2015,guigues_convergence_2016,dowson_policy_2018}. Because SDDP uses convex optimization, the algorithm can find policies for problems with thousands of continuous control variables and a complicated, but still convex, feasible \mbox{region}.}


\revision{Our decision-dependent one-step transition probabilities introduce non-convexities. {The convexity assumptions of SDDP have been interpreted as precluding its application to non-convex problems. However, a recent trend computes a convex value function approximation to define a sub-optimal policy, which is then evaluated using the original non-convex model.} Our work is part of this trend. When non-convexities arise due to discrete variables, Lagrangian duality can construct a convex approximation \cite{zouStochasticDualDynamic2019}. This technique has been applied, for example, to infrastructure planning problems in electric power systems \cite{laraEJOR2018,laraSDDiP2020}. In hydrothermal scheduling, Rosemberg et al.\  \cite{rosemeberg2022} build a value function {using convex} approximations of the alternating current optimal power flow (AC-OPF) problem, and then evaluate the resulting policy using non-convex AC-OPF dynamics. In a problem {with} a hidden Markov model, Siddig et al.\  \cite{siddig2021maximum} convexify the value function by assuming a fully observable Markov chain, but then evaluate the policy on the hidden model.}


\revision{While inherent in MDPs, decision-dependent uncertainty, in which the agent's actions change the probability distribution, is less common in stochastic programs. That said, there is a stream of stochastic programming literature on this topic, and we point to  \cite[Chapter 5]{adiga2023} and \cite{hellemo2018decision} for reviews. In one class of problems, the agent's decisions alter the probability mass function of a fixed set of realizations according to a decision-dependent function \cite{arslan2024sddp,lejeune2018chance,yin2022coordinated}. In a second class of problems, the points of support, or in time-dynamic problems the information structure, is altered by the agent's actions \cite{basciftci2024adaptive,goel2006class,lamas2024target}. In isolation, our decision-dependent approach is of the former type, although we alter one-step transition probabilities in a Markov chain rather than the probability mass of a random vector, and our approach applies to multi-stage problems with an infinite horizon. When combined with statistical learning, our approach can be of the second type in that the agent's actions can alter the resulting points of support.}


{The paper focuses} on demonstrating the flexibility and expressiveness of {our approach} to modeling---primarily through several examples---although we derive requisite new SDDP algorithms. {The} contributions of our paper are:
\begin{enumerate}
    \item We extend policy graphs and SDDP to allow the one-step transition probabilities to depend on our decisions.
    \item We incorporate learning so an agent can probe and gather information to improve their beliefs before committing to a decision or, more generally, while committing to a sequence of decisions. 
\end{enumerate}

\revision{From one lens, these contributions add MDP- and RL-style modeling features to MSPs. From another, our SDDP algorithms can approximately solve a class of CACS MDPs.} {To summarize, we extend SDDP to handle a class of stochastic programs defined on Markov chains that incorporate decision-dependent uncertainty and learning, enabling continuous-state MDP modeling without discretization.}

{The paper} is laid out as follows. Section~\ref{sec:policy_graphs} introduces policy graphs, which we extend to decision-dependent transitions, to a form of statistical learning, and to decision-dependent learning in Sections~\ref{sec:modeling-ddu}, \ref{sec:learning}, and~\ref{sec:dd_learning}. Section~\ref{sec:sddp} summarizes a basic SDDP algorithm, which we extend to decision-dependent learning in Sections~\ref{sec:ddu-sddp} and~\ref{sec:SDDP_for_dd_learning}.  Section~\ref{sec:computational-results} presents numerical experiments. 

\section{Policy graphs for MDPs}\label{sec:policy_graphs}

With an eye towards extensions, we present---with some necessary changes to the notation and definitions---the \textit{policy graph} \revision{modeling} framework of \cite{dowson_policy_2018}. We use $x$ for the physical state variable and $u$ for the control variable, consistent with the stochastic control literature. The goal in this section is to establish the core modeling approach and then extend its reach in subsequent sections.

\subsection{Mathematical model}\label{sec:policy_graphs_math_models}

\begin{definition}\label{def:policy_graph}
A \textit{policy graph}, $\mathcal{G}=(R,\mathcal{N},\Phi)$, is composed of a tuple with a root node $R$, a further set of nodes $\mathcal{N}$, and an $|\mathcal{N}|+1$ by $|\mathcal{N}|$ matrix $\Phi$ that specifies one-step transition probabilities, with entries $\phi_{ij}$. We use $\Phi_0$ to denote the square matrix of transition probabilities on $\mathcal{N}$, dropping the root. The {\it children} of node $i$ are $i^+ = \{j \in \mathcal{N}: \phi_{ij} > 0\}$. A policy graph \revision{specifies a discrete-time,} time-homogeneous, absorbing Markov chain \revision{with} initial state specified by $\phi_{R,i}$, transition probabilities among transient states given by $\Phi_0$, and \revision{a single absorbing state}. \revision{For each $i$, $1 - \sum_{j\in i^+} \phi_{ij} \ge 0$ is the one-step probability of transitioning to the} absorbing state, which \revision{has cost zero} and ends a sequence of transitions. We do not explicitly model the absorbing state.

\revision{A policy graph includes} a \textit{decision rule}, $\pi_i(x, \omega_i)$, for each node that maps the \textit{incoming \revision{physical} state variable} $x$ and realization of a \textit{random variable}~$\omega_i$ to a feasible \textit{control variable} $u$ and an \textit{outgoing \revision{physical} state variable} $x^\prime$ for a cost of $C_i( u, x^\prime, \omega_i)$, where $(u, x^\prime) =\pi_i(x, \omega_i) \in \mathcal{X}_i(x, \omega_i) \equiv \{ (u, x^\prime): u \in U_i(x, \omega_i), \ x^\prime = T_i(u, x, \omega_i) \revision{\in {\mathcal X}'_i} \}$. \revision{The} random variable $\omega_i$ is independent of $x$ and of $\omega_j$ at all other nodes, and has a finite sample space $\Omega_i$ with probability mass function~(pmf) $\mathbb{P}_i(\omega)$,  $\omega \in \Omega_i$.  \qed 
\end{definition}

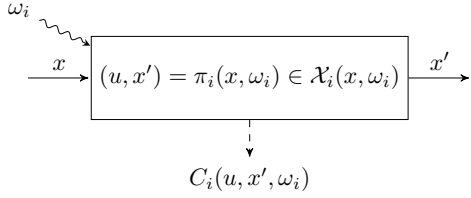
\begin{figure}[ht]
    \centering
\scalebox{0.90}{    
    \begin{tikzpicture}[align=center,node distance=3.4cm]
        \node[textnode] (a)  [] {};
        \node[square] (b)  [right of = a] {\color{white}{a}\\$(u, x^\prime) =\pi_i(x, \omega_i) \in \mathcal{X}_i(x, \omega_i)$\\};
        \node[textnode] (w) at (0,1) {$\omega_i$};
        \node[textnode] (corner) at (1.2,0.4) {};
        \node[textnode] (cst)  at (3.4, -1.5) {$C_i( u, x^\prime, \omega_i)$};
        \node[textnode] (c)  [right of = b] {};
        \path[]
        	(a)	edge 	[] 	node	[left] 	[above] {$x$}	(b)
        	(b)	edge 	[dashed] 	node	[left] 	[above] {}	(cst)
            (b)	edge 	[] 	node	[left] 	[above] {$x^\prime$}	(c);
        \draw[-stealth, decoration={snake, amplitude = .4mm, segment length = 2mm, post length=0.9mm},decorate] (w) -- (corner);
        \node [align=center] (ww) at (0.8,0.8) {};
    \end{tikzpicture}
}    
    \caption{\revision{A schematic of a} node in a policy graph. The incoming \revision{physical} state is denoted by $x$ and an incoming realization by $\omega_i$. The \revision{decision rule} $\pi_i(x,\omega_i)$ specifies the control, $u$, and outgoing \revision{physical} state,~$x'$. We incur a one-step cost $C_i( u, x^\prime, \omega_i)$, and transition according to one-step probabilities $\phi_{ij}$ (not shown) to a child node, or terminate with probability $1 - \sum_{j \in i^+} \phi_{ij}$.}
    \label{fig:hd}
\end{figure}

\revision{
Using the concept and notation of a policy graph, we can now define the optimization problem that we study in the remainder of this paper. In what follows, we assume that a common space ${\mathcal X}'$ contains ${\mathcal X}'_i$ for all $i \in {\mathcal N}$, where ${\mathcal X}'$ is compact.
\begin{definition}\label{defn:policy_graph}
A risk-neutral optimization problem on a policy graph is
\begin{equation}\label{eqn:full_problem_without_super_pi}
\min_{\pi \in \Pi} \E_{i \in R^+; \, \omega_i}[V_{1,i}(x_R, \omega_i)],
\end{equation}
where for $i \in {\mathcal N}$ and $t=1,2,\ldots$
\begin{equation*}
V_{t,i}(x, \omega_i) = \left \{ C_i( u, x^\prime, \omega_i) + \E\limits_{j\in i^{+};\, \omega_j} \left[V_{t+1,j}(x^\prime, \omega_j)\right] \right \},
\end{equation*}
and where the control and outgoing physical state variables are specified by policy $\pi$'s non-anticipative decision rules, which satisfy:
\begin{equation*}
(u, x^\prime) = \pi_i(x, \omega_i) \in \mathcal{X}_i(x, \omega_i). \tag*{\qed}
\end{equation*}   
\end{definition}}
{In Definition~\ref{defn:policy_graph}, $t$ counts transitions from the root along a sample path, and $i$ indexes the Markov chain's current node. In the Bellman reformulation of Section~\ref{section:Bellman_reformulation}, the value function will depend only on $i$, with $t$ dropping out.} 

 \revision{A solution to problem~\eqref{eqn:full_problem_without_super_pi} is a policy $\pi^*$ that minimizes the expected cost, starting from the root. The expectation operator, $\E[ \cdot ]$, accounts for transition probabilities $\phi_{ij}$ from node $i$ to nodes $j \in i^+$, and the random variable $\omega_j$:
$$\E\limits_{j\in i^{+};\, \omega_j} \left[V_{t+1,j}(x^\prime, \omega_j)\right] = \sum\limits_{j \in i^+} \phi_{ij} \sum\limits_{\omega \in \Omega_j} \mathbb{P}_j(\omega) \cdot V_{t+1,j}(x^\prime, \omega).$$
When summing, or otherwise enumerating over $\omega \in \Omega_j$, we typically suppress $\omega$'s superfluous index, $j$, unless necessary to disambiguate.}

\revision{If a policy graph is acyclic (see Example~\ref{example:two-stage} below), the problem has a finite horizon. If $i$ is a leaf node of ${\mathcal G}$ then $i^+ = \emptyset$, and so we need not specify a terminal value function.  If a policy graph contains cycles (see Example~\ref{example:two-stage-cyclic} below), the problem has an infinite horizon. In such problems, one can minimize long-run average cost or total discounted cost. Our approach is of the latter type because the nodes $i \in {\mathcal N}$ are transient states in an absorbing Markov chain. Given an ergodic chain and a discount factor (e.g., $\rho =0.95$) that applies at each time step, in order to use our formulation, we multiply each entry of the one-step transition matrix by $\rho$ so that $\Phi_0$ would become a sub-stochastic matrix for transitions among transient states of an absorbing chain.} 

\revision{When modeling, the information sufficient to communicate the formulation of problem~\eqref{eqn:full_problem_without_super_pi} is: the set of nodes $\mathcal{N}$; the transition matrix $\Phi$; the physical state at the root $x_R$; the random variable $\omega_i$ with pmf $\mathbb{P}_i$ for all $i \in \mathcal{N}$; and the \textit{subproblems}, $\mathbf{SP}_i(x, \omega_i)$, defined as the constrained optimization problem:
\begin{equation*}
\mathbf{SP}_i(x, \omega_i): \ \  \ \min\limits_{u, x^\prime}\left\{ C_i( u, x^\prime, \omega_i) \;\big\vert\; (u, x^\prime) \in \mathcal{X}_i(x, \omega_i)\right\}.
\end{equation*}
}

\subsection{\revision{Bellman recursion reformulation}}\label{section:Bellman_reformulation}

\revision{
Problem~\eqref{eqn:full_problem_without_super_pi} can be reformulated using Bellman's equation \cite{bellman_dynamic_1957}:
\begin{equation}\label{eq:DH}
\begin{array}{r l}
    V_i(x, \omega_i) = \min\limits_{u, x^\prime}& \left \{ C_i( u, x^\prime, \omega_i) + \E\limits_{j\in i^{+};\, \omega_j} \left[V_j(x^\prime, \omega_j)\right] \right \} \\
  \text{s.t.} & (u, x^\prime) \in \mathcal{X}_i(x, \omega_i),
\end{array}
\end{equation}
and an optimal policy~$\pi^*$ can be obtained by choosing a control $u$ in the argmin of problem~\eqref{eq:DH}:
$$\pi_i(x, \omega_i) \in \argmin\limits_{(u, x^\prime) \in \mathcal{X}_i(x, \omega_i)} \left\{C_i( u, x^\prime, \omega_i) + \E\limits_{j\in i^{+};\, \omega_j} \left[V_j\left(x^\prime, \omega_j \right)\right]\right\}.$$
}

\revision{ For an acyclic policy graph, i.e., a finite-horizon problem, the mapping between formulations~\eqref{eqn:full_problem_without_super_pi} and~\eqref{eq:DH} is straightforward. Unrolling recursion~\eqref{eq:DH} on an acyclic graph allows $i$ in~\eqref{eq:DH} to encode $i$ and $t$ in~\eqref{eqn:full_problem_without_super_pi}. {(See discussion in subsequent Example~\ref{example:two-stage-cyclic}.)} For a cyclic policy graph, i.e., an infinite-horizon problem, the reformulation is justified by considering the Bellman operator ${\mathcal B}$:
\begin{equation}\label{eqn:bellman_operator}
\begin{array}{r l}
    {\mathcal B} v_i(x, \omega_i) = \min\limits_{u, x^\prime}& \left \{ C_i( u, x^\prime, \omega_i) + \E\limits_{j\in i^{+};\, \omega_j} \left[v_j(x^\prime, \omega_j)\right] \right \} \\
  \text{s.t.} & (u, x^\prime) \in \mathcal{X}_i(x, \omega_i),
\end{array}
\end{equation}
which applies to the space of bounded functions, $v$, on ${\mathcal D} \equiv \cup_{i \in \mathcal{N}} ({\mathcal X}' \times \{i\} \times \Omega_i)$. Let
$\| v \|_\infty = \sup_{(x,i,\omega) \in {\mathcal D}} |v_i(x,\omega)|$,
and let $\gamma \in (0,1)$. We say that  ${\mathcal B}$ is a $\gamma$-contraction if for any pair of bounded functions $v$ and $v'$ on domain ${\mathcal D}$ we have $\| {\mathcal B} v - {\mathcal B} v' \|_\infty \le \gamma || v - v' ||_\infty$.}

\revision{
Writing the fixed-point condition ${\mathcal B} V=V$ is synonymous with equation~\eqref{eq:DH} holding for all $(x,i,\omega_i) \in {\mathcal D}$.  If ${\mathcal B}$ a $\gamma$-contraction, then ${\mathcal B} V=V$ has a unique fixed point, and solving the infinite horizon problem~\eqref{eqn:full_problem_without_super_pi} is equivalent to solving the fixed-point equation~\eqref{eq:DH}. In this case, the Bellman policy is optimal, so that:
\begin{equation*}
\min_{\pi \in \Pi} \E_{i \in R^+; \, \omega_i}[V_{t,i}(x_R, \omega_i)] = \E_{i \in R^+; \, \omega_i}[V_i(x_R, \omega_i)].
\end{equation*}
}

\revision{Although our Bellman reformulation of a cyclic policy graph problem~\eqref{eqn:full_problem_without_super_pi} hinges on a value-function argument, the policy graph is a \textit{modeling framework} that is independent of the solution method. In this paper, we compute policies using   SDDP-style algorithms, i.e., value-function approximation algorithms, but given a policy graph, one could apply other algorithms. 
}

\subsection{Assumptions}\label{sec:assumptions}

\revision{
To \revision{ensure our formulation is well-defined} and to facilitate computationally tractable algorithms, we make the following assumptions that restrict the space of policy graphs that we consider:
\begin{enumerate}
    \item[(A1)] The number of nodes, $\mathcal{N}$, is finite.
    \item[(A2)] The sample space, $\Omega_i$, is finite at each node $i \in \mathcal{N}$, and the random variables $\omega_i$, $i \in \mathcal{N}$, are mutually independent. 
    \item[(A3)] For $i \in \mathcal{N}$, $\mathcal{X}_i(x, \omega_i) = \{ (u, x^\prime): u \in U_i(x, \omega_i), \ x^\prime = T_i(u, x, \omega_i) \revision{\in {\mathcal X}'_i} \} \neq \emptyset$ and $U_i(x, \omega_i)$ is compact $\forall x \in {\mathcal X}'$, $\omega_i \in \Omega_i$, where ${\mathcal X}'_i \subseteq {\mathcal X}'$ and where ${\mathcal X}'$ is compact.
    \item[(A4)] $\min\limits_{u, x^\prime, x}\left\{ C_i( u, x^\prime, \omega_i) \;\big\vert\;  (u, x, x^\prime) \in \mathcal{X}_i(\omega_i) \equiv \{ (u, x, x^\prime) \, : \, (u, x^\prime) \in \mathcal{X}_i(x,\omega_i) \} \right\}$ is a convex optimization model for each $i \in \mathcal{N}$ and $\omega_i \in \Omega_i$. 
    \item[(A5)] The one-step matrix $\Phi_0$ is such that either the policy graph is acyclic or the Bellman operator~${\mathcal B}$ in equation~\eqref{eqn:bellman_operator} is a $\gamma$-contraction. 
\end{enumerate}
In addition to (A1) and (A2), the algorithms we develop in this paper require that $|\Omega_i|$ and $|\mathcal{N}|$ are of modest size, e.g., at most 100 (to an order of magnitude). 
Under (A3)---which includes relatively complete recourse---subproblem $\mathbf{SP}_i(x, \omega_i)$ is feasible, and the subproblem further has a finite optimal solution because its objective function is convex by (A4). Assumption (A4)~further ensures that problem~\eqref{eqn:full_problem_without_super_pi} is convex with respect to $x$ and $u$. 
As we discuss in Section~\ref{section:Bellman_reformulation}, (A5) ensures that the Bellman reformulation is equivalent to Section~\ref{sec:policy_graphs_math_models}'s model. That said, in the infinite-horizon setting, we want to be able to verify that (A5) holds in terms of our problem's primitives. Given our assumptions, including compactness of $\mathcal{U}_i(x,\omega_i)$ and $\mathcal{X}'$, a sufficient condition on the one-step transition probabilities, $\Phi_0$, to ensure (A5) holds is that $\gamma \equiv \max_{i \in \mathcal N} \sum_{j \in i^+} \phi_{ij} < 1$. In this case, a standard argument shows that ${\mathcal B}$ is a $\gamma$-contraction.   
}

\subsection{{Policy graphs and MDPs}}

\revision{Section~\ref{sec:introduction} denotes the state by $s \in \mathcal{S}$, which is the pervasive convention in discrete-action discrete-state MDPs. The analogous state in a policy graph for our CACS MDP is $(x,i)$, where $x$ is a vector of continuous physical state variables (we allow some discrete variables in what follows) and~$i$ is one of a finite number of Markov-chain states. As indicated, $(u, x^\prime) \in \mathcal{X}_i(x, \omega_i)$ is shorthand for $u \in U_i(x, \omega_i)$ and $x^\prime = T_i(u, x, \omega_i) \in \mathcal{X}'_i$. Thus $U_i(x,\omega_i)$ and $\mathcal{X}'_i$ constrain the control---the analog of $a \in \mathcal{A}_s$ in our typical MDP notation---and given $u$, $T_i(u, x, \omega_i)$ determines the outgoing state. In this way, $x^\prime=T_i(u, x, \omega_i)$ specifies a (conditionally) deterministic} transition for the $x$-component of the state, and our one-step transition matrix $\Phi$ specifies a probabilistic transition for the $i$-component of the state. Section~\ref{sec:introduction} points to a distribution, $\mu(s)$, on initial states. Consistent with the transitions in a policy graph just mentioned, we initialize the root node with a deterministic physical state, $x_R$, but allow a random transition from the root, $\phi_{R,i}$, which is the analog of $\mu(s)$ but only for the $i$-component of the state. 

{Computation with MDPs is typically accomplished using discrete action and state spaces. A policy graph, combined with SDDP, offers an alternative. Exploiting the convexity assumptions of Section~\ref{sec:assumptions}, we can handle continuous states and actions directly, constructing a piecewise linear outer approximation of the cost-to-go function without discretization.}



\subsection{{Policy graphs and MSPs}}

{A multi-stage stochastic program with a finite horizon, finite set of scenarios, and general inter-stage dependence is typically defined on a scenario tree. Such a model is a special sub-class of policy graphs in which the matrix $\Phi$ defines a tree and the random variable at each node is degenerate, i.e., $|\Omega_i| = 1$. Here, the nodes and arcs of the scenario tree correspond exactly to the nodes and arcs of the equivalent policy graph. If the MSP is inter-stage independent it could again be represented as just sketched, but with $\Phi$ and the degenerate realizations at each node encoding inter-stage independence. That said, policy graphs afford an attractive and compact alternative, illustrated in Figure~\ref{example-fig:standard_T_stage_MSP}.} 

\begin{figure}[ht]
    \centering
\scalebox{0.90}{
    \begin{tikzpicture}[align=center]
        \node[bellman] (a) at (0, 0) {$R$};
        \node[square] (b) at (2, 0) {1};
        \node[textnode] (b_w) at (1.5, 1) {};
        \node[square] (c) at (4, 0) {$2$};
        \node[textnode] (c_w) at (3.5, 1) {};
        \node[textnode] (d) at (6, 0) {...};
        \node[square] (e) at (8.5, 0) {$T$};
        \node[textnode] (e_w) at (8, 1) {};
        \path[]
        	(a)	edge 	[] 	node	[left] 	[above] {$\phi_{R,1}=1$}	(b)
            (b)	edge 	[] 	node	[left] 	[above] {$\phi_{1,2}=1$}	(c)
            (c)	edge 	[] 	node	[left] 	[above] {$\phi_{2,3}=1$}	(d)
            (d)	edge 	[] 	node	[left] 	[above] {$\phi_{T-1,T}=1$}	(e);
        \drawsquiggle{c_w}{c};
        \drawsquiggle{e_w}{e};
    \end{tikzpicture}
}
    \caption{{Policy graph structure for a standard $T$-stage MSP with inter-stage independence. We refer to this as a {\it linear} policy graph.}}
    \label{example-fig:standard_T_stage_MSP}
\end{figure}

{In this way, all problems defined on a finite scenario tree can be captured using policy graphs, but the converse is false. Figure~\ref{figure:markov_switching_four_quarters} depicts what would be a four-stage stochastic program, except that return arcs with probability $\rho < 1$ yield an infinite-horizon problem. This example is defined using an (infinite-horizon) Markov process rather than a finite scenario tree. The next subsection begins to develop such examples in more detail. 
}

\begin{figure}[ht]
    \centering
\scalebox{0.90}{
\begin{tikzpicture}[scale=2.0,
  node/.style={rectangle, draw, minimum size=3.5mm, inner sep=0pt},
  circnode/.style={circle, draw, minimum size=7mm, inner sep=0pt},
  arrow/.style={draw, -{Latex[length=1.4mm,width=1.2mm]}},
  squigarrow/.style={
    draw,
    decorate,
    decoration={snake, amplitude=0.35mm, segment length=1.5mm},
    -{Latex[length=1.4mm,width=1.2mm]}
  }
]

  \node[node] (a1) {};
  \node[node] (b1) [right=1cm of a1] {};
  \node[node] (c1) [right=1cm of b1] {};
  \node[node] (d1) [right=1cm of c1] {};

  \node[node] (a2) [below=7mm of a1] {};
  \node[node] (b2) [right=1cm of a2] {};
  \node[node] (c2) [right=1cm of b2] {};
  \node[node] (d2) [right=1cm of c2] {};

\node[circnode] (R) at ($(a1.west)!0.5!(a2.west) + (-0.66cm,0)$) {$R$};

  \path
    (R) edge[arrow] node[midway, above, font=\small] {$\tfrac12$} (a1)
    (R) edge[arrow] node[midway, below, font=\small] {$\tfrac12$} (a2);

  \path
    (a1) edge[arrow] (b1)
    (b1) edge[arrow] (c1)
    (c1) edge[arrow] (d1)
    (a2) edge[arrow] (b2)
    (b2) edge[arrow] (c2)
    (c2) edge[arrow] (d2);

  \path
    (a1) edge[arrow] (b2)
    (a2) edge[arrow] (b1)
    (b1) edge[arrow] (c2)
    (b2) edge[arrow] (c1)
    (c1) edge[arrow] (d2)
    (c2) edge[arrow] (d1);

\draw[arrow]
  (d1.north east)
    to[out=45, in=135, looseness=1.2]
    node[midway, above, font=\small] {$\rho$}
  (a1.north);

\draw[arrow]
  (d2.south east)
    to[out=-45, in=-135, looseness=1.2]
    node[midway, below, font=\small] {$\rho$}
  (a2.south);

  \foreach \v in {a1,b1,c1,d1}{
    \draw[squigarrow]
      ([xshift=-2.5mm,yshift=2.5mm]\v.north west) -- (\v.north west);
  }
  \foreach \v in {a2,b2,c2,d2}{
    \draw[squigarrow]
      ([xshift=-2.5mm,yshift=-2.5mm]\v.south west) -- (\v.south west);
  }
\end{tikzpicture}
}
\caption{{In each fiscal quarter, we can be in a volatile or normal state (top and bottom rows). Each is equally likely initially, and there is Markov switching each quarter, meaning the agent faces a mixture model. Further transition probabilities are suppressed except those that return from the final quarter of one year to the first quarter of the next, with probability $\rho < 1$. (For simplicity we do not show switching for this return.) This is a cyclic policy graph.}\label{figure:markov_switching_four_quarters}}
\end{figure}

{A policy graph with a node $i$ for which $|\Omega_i| > 1$ can be expanded into a larger policy graph with an extra node for each realization of $\omega_i$. Conversely, two nodes in a policy graph can be combined if they share identical expected cost-to-go functions and identical functional forms of the constraint set; see~\eqref{eq:DH}. A computational advantage of a parsimonious policy graph, i.e., one with fewer nodes, is that SDDP-style algorithms can then exploit shared expected cost-to-go functions.} 


\subsection{Examples}\label{sec:basic-examples}

We provide examples that begin to demonstrate the modeling flexibility of policy graphs.
\revision{For clarity, we present minimal instances that isolate essential elements of our approach. These examples are not merely illustrative. Rather, they are canonical building blocks that recur as sub-structures in large-scale models, and the modeling insights developed here extend to those settings.}

\begin{example}[Two-stage newsvendor]\label{example:two-stage}
Consider a two-stage newsvendor problem. In the first stage, the agent decides the number of newspapers to order, which cost \$2 each. In the second stage, the agent observes demand~$\omega$ and can sell newspapers at a unit price of \$5. Excess newspapers must be disposed of at a unit cost of \$0.1. Figure~\ref{example-fig:two-stage} shows the graph structure and subproblems. {The problem can also be formulated as shown in Figure~\ref{example-fig:two-stage-scenario-tree}, demonstrating how scenario trees from stochastic programming are a special case of a policy graph.}

\begin{figure}[ht]
    \centering
\scalebox{0.90}{     
    \begin{tikzpicture}[align=center]
        \node[bellman] (a) at (0, 0) {$x_R = 0$};
        \node[square] (b)  at (4.5, 0) {
            $\begin{array}{r l}
                \min\limits_{u, x^\prime}& 2u \\
              \text{s.t.} & x^\prime = x + u \\
                          & u \ge 0
            \end{array}$};
        \node[square] (b2)  at (10, 0) {
            $\begin{array}{r l}
                \min\limits_{u, x^\prime}& -5u + 0.1x^\prime \\
              \text{s.t.} & u \le x \\
                          & x^\prime = x - u \\
                          & 0 \le u \le \omega
            \end{array}$};
        \node[textnode] (w2) at (7.0,1.5) {$\omega$};
        \path[]
        	(a)	edge 	[] 	node	[left] 	[above] {$\phi_{R,1}=1$}	(b)
            (b)	edge 	[] 	node	[left] 	[above] {$\phi_{1,2}=1$}	(b2);
        \draw[-stealth, decoration={snake, amplitude = .4mm, segment length = 3mm, post length=1mm},decorate] (w2) -- (b2);
        \node [align=center] (ww) at (6.4,1.3) {};     
    \end{tikzpicture}
}
    \caption{The policy graph for Example~\ref{example:two-stage}. While the problem is formulated with continuous actions and physical state, in some problem variants we instead require integer-valued decisions, e.g., $u \in \mathbb{Z}_+$ at node 1 and $u \in \{0, 1, \ldots, \omega\}$ at node 2.}
    \label{example-fig:two-stage}
\end{figure}

\begin{figure}[ht]
    \centering
\scalebox{0.9}{     
    {\begin{tikzpicture}[align=center]
        \node[bellman] (a) at (0, 0) {$x_R = 0$};
        \node[square] (b)  at (4.5, 0) {
            $\begin{array}{r l}
                \min\limits_{u, x^\prime}& 2u \\
              \text{s.t.} & x^\prime = x + u \\
                          & u \ge 0
            \end{array}$};
        \node[square] (b2)  at (11.5, -2.0) {
            $\begin{array}{r l}
                \min & -5u + 0.1x^\prime \\
              \text{s.t.} & 0 \le u \le \min\{\omega_3, x\} \\
                          & x^\prime = x - u
            \end{array}$};
        \node[square] (b3)  at (11.5, 0) {
            $\begin{array}{r l}
                \min & -5u + 0.1x^\prime \\
              \text{s.t.} & 0 \le u \le \min\{\omega_2, x\} \\
                          & x^\prime = x - u
            \end{array}$};
        \node[square] (b4)  at (11.5, 2.0) {
            $\begin{array}{r l}
                \min & -5u + 0.1x^\prime \\
              \text{s.t.} & 0 \le u \le \min\{\omega_1, x\} \\
                          & x^\prime = x - u
            \end{array}$};
        \path[]
        	(a)	edge 	[] 	node	[left] 	[above] {$\phi_{R,1}=1$}	(b)
            (b)	edge 	[] 	node	[left] 	[below] {$\phi_{1,4}=\mathbb{P}_2(\omega_3)$}	(b2)
            (b)	edge 	[] 	node	[left] 	[above] {$\phi_{1,3}=\mathbb{P}_2(\omega_2)$}	(b3)
            (b)	edge 	[] 	node	[] 	[above] {$\phi_{1,2}=\mathbb{P}_2(\omega_1)$}	(b4);
    \end{tikzpicture}}
}
    \caption{{The policy graph for Example~\ref{example:two-stage}, expanded into a scenario tree with $\Omega_2 = \{\omega_1, \omega_2, \omega_3\}$. Note that the $\omega_i$ in the second-stage problems are constants, i.e., degenerate random variables, and their corresponding probability masses are now on the arcs.}}
    \label{example-fig:two-stage-scenario-tree}
\end{figure}
\end{example}

\begin{example}[Cyclic newsvendor]\label{example:two-stage-cyclic}
We can extend Example~\ref{example:two-stage} to an infinite-horizon by adding a cyclic second-stage subproblem, which includes both buying $u_b$ and selling $u_s$. The disposal cost of \$0.1 is now a holding cost. Figure~\ref{example-fig:two-stage-cyclic} shows the graph structure and subproblems.  We can form a finite-horizon approximation by ``unrolling'' the loop to form {a $T$-stage problem as in Figure~\ref{example-fig:standard_T_stage_MSP}, except that $\phi_{2,3}=\cdots=\phi_{T-1,T}=\rho$}.

\begin{figure}[ht]
    \centering
\scalebox{0.90}{     
    \begin{tikzpicture}[align=center]
        \node[bellman] (a) at (0, 0) {$x_R = 0$};
        \node[square] (b)  at (4.5, 0) {
            $\begin{array}{r l}
                \min\limits_{u, x^\prime}& 2u \\
              \text{s.t.} & x^\prime = x + u \\
                          & u \ge 0
            \end{array}$};
        \node[square] (b2)  at (10, 0) {
            $\begin{array}{r l}
                \min\limits_{u, x^\prime}& -5u_s + 2u_b + 0.1x^\prime \\
              \text{s.t.} & u_s \le x \\
                          & x^\prime = x - u_s + u_b \\
                          & 0 \le u_s \le \omega \\
                          & u_b \ge 0
            \end{array}$};
        \node[textnode] (w2) at (6.5,1.5) {$\omega$};
        \path[]
        	(a)	edge 	[] 	node	[left] 	[above] {$\phi_{R,1}=1$}	(b)
            (b)	edge 	[] 	node	[left] 	[above] {$\phi_{1,2}=1$}	(b2);
        \draw[-stealth, decoration={snake, amplitude = .4mm, segment length = 2mm, post length=1mm},decorate] (w2) -- (b2);
        \node [align=center] (ww) at (6.4,1.3) {};
        \draw (b2) to [out=-60,in=-120,looseness=3] node {$\phi_{2,2} = \rho$}(b2);    
    \end{tikzpicture}
}
    \caption{The policy graph for Example~\ref{example:two-stage-cyclic}. The same comments from Example~\ref{example:two-stage} hold regarding continuity of~$u$.}\label{example-fig:two-stage-cyclic}
\end{figure}
\end{example}

\begin{example}[Markovian  newsvendor]\label{example:markovian}
Suppose that there are two market states, sunny and cloudy, and transitions between them over time can be modeled as a Markov chain. When the weather is sunny, demand is governed by a favorable random variable $\omega_s$, and when it's cloudy, the random demand is instead $\omega_c$. The subproblems are the same as in node 2 of Example~\ref{example:two-stage-cyclic}, and the root state is again $x_R = 0$. The graph structure is shown in Figure~\ref{example-fig:markovian}.
\begin{figure}[ht]
    \centering
    \begin{tikzpicture}[align=center]
        \node[bellman] (a) at (2, 0) {$R$};
        \node[square] (h)  at (4, 0.7) {$s$};
        \node[square] (l)  at (4, -0.7) {$c$};
        \node[textnode] (w_h) at (3.5,1.5) {$\omega_s$};
        \node[textnode] (w_l) at (3.5,0.1) {$\omega_c$};
        \path[]
        	(a)	edge 	[] 	node	[left] 	[above] {}	(h)
            (a)	edge 	[] 	node	[left] 	[below] {}	(l);
        \draw[-stealth, decoration={snake, amplitude = .4mm, segment length = 2mm, post length=0.9mm},decorate] (w_h) -- (h);
        \draw[-stealth, decoration={snake, amplitude = .4mm, segment length = 2mm, post length=0.9mm},decorate] (w_l) -- (l);
        \draw (l) to [out=100,in=-100] node {}(h);
        \draw (l) to [out=0,in=-90,looseness=8] node {}(l);
        \draw (h) to [out=0,in=90,looseness=8] node {}(h);
        \draw (h) to [out=-80,in=80] node {}(l);
    \end{tikzpicture}
    \caption{The policy graph for Example~\ref{example:markovian}. When the process exits the sunny node, it returns with probability $\phi_{s,s}$, transitions to the cloudy node with probability $\phi_{s,c}$, and the process terminates with probability $1-\phi_{s,s}-\phi_{s,c}$. Analogous probabilistic transitions occur out of the cloudy market state.}
    \label{example-fig:markovian}
\end{figure}
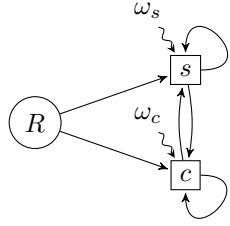
\end{example}

\section{Decision-dependent transitions}\label{sec:modeling-ddu}

\subsection{Mathematical model}\label{sec:decision_dep_math_models}

In an MDP, the one-step transition probabilities, $\mathbb{P}(s^\prime | s, a)$, depend on the agent's action. In our policy graph formulation, the next state $(x^\prime,\revision{j})$, depends on $x'=T_i(u, x, \omega_i)$ and $\phi_{i\revision{j}}$. The former deterministic transition depends on the agent's action, but the latter probabilistic transition does not. 
Restated, our formulation of Section~\ref{sec:policy_graphs_math_models} has one-step transition probabilities, $\Phi$, which do not depend on the agent's decisions.  

In this section we extend policy graphs to allow action-dependent matrices, $\Phi(y)$, where $y$ is an additional decision variable. In the stochastic programming literature, this is one form of \revision{decision-dependent uncertainty \cite{ahmed2000strategic,arslan2024sddp,dupacova2006optimization,hellemo2018decision}.}

We give two formulations to handle action-dependent one-step transition matrices. Each formulation lends itself to a different SDDP variant, which we discuss in Section~\ref{sec:sddp_whole_section}. In our first decision-dependent model we have:
\begin{equation*}
\min_{\pi \in \Pi} \sum_{i \in R^+}  \phi_{R,i} \sum_{\omega \in\Omega_i}\mathbb{P}_i(\omega)\cdot V_{i}(x_R,\omega),
\end{equation*}
where:
{\scalefont{0.95}
\begin{equation}\label{eq:dec_dep_formulation_y}
\begin{array}{r l}
    V_i(x, \omega_i) =
    \min\limits_{u, x^\prime, y}& C_i( u, x^\prime, \omega_i) + C^\prime_i(y) + \sum\limits_{j\in i^{+};\, \omega \in\Omega_j} \phi_{ij}(y) \cdot \mathbb{P}_j(\omega)\cdot V_j(x^\prime, \omega) \\
  \text{s.t.} &(u, x^\prime) \in \mathcal{X}_i(x, \omega_i) \\
  & y \in \mathcal{Y}.
\end{array}
\end{equation}
}

\noindent Decision $y$ could represent marketing, which increases the likelihood of a favorable demand distribution. Such a decision incurs a cost $C^\prime_i(y)$, which we assume is linear and independent of the control decisions. We assume $0 \in \mathcal{Y}$, where $\mathcal{Y}$ is a polytope, and for simplicity, we assume identical constraints $y \in \mathcal{Y}$ at each node $i$. In a more general model, $y$ could compete with $u$ for limited resources and the set $\mathcal{Y}$ may differ between nodes. For the one-step transition matrices, we assume a linear response to $y$:
\begin{equation}\label{eqn:linear_marketing}
\phi_{ij}(y) = \phi_{ij}^{0} ( 1 + a_{j}^\top y ), \, \mbox{for} \, i, j \in \mathcal{N},
\end{equation}
where $\phi_{ij}^{0}$ is a nominal one-step transition probability when $y=0$. We then require $( 1 + a_{j}^\top y ) \ge 0, \ \forall y \in \mathcal{Y}$, and $\sum_{j \in i^+} \phi_{ij}^{0} a_{j} = 0$. Because $C^\prime_i(y)$ and $\phi_{ij}(y)$ are both linear in $y$, we can restrict attention to the extreme points of $\mathcal{Y}$ in the minimization of~\eqref{eq:dec_dep_formulation_y}. In what follows we will assume $\mathcal{Y}$ is finite with a modest number of options at each node in the policy graph. Even in this setting, there are a huge number of such options in the overall problem. 

Our second formulation is motivated by the above development but allows greater generality. We assume a limited set of possible transition matrices $\{\Phi^d\}_{d\in D}$, and at each node, a binary variable $y_d \in \{0, 1\}$ specifies the matrix for the next step. The cost-to-go function at node $i\in\mathcal{N}$ is now:
{\scalefont{0.95}
\begin{equation}\label{eq:dec_dep_formulation_y_discrete}
\begin{array}{r l}
    V_i(x, \omega_i) =
    \min\limits_{u, x^\prime, y}& C_i(u, x^\prime, y, \omega_i) + \sum\limits_{d \in D} y_d \sum\limits_{j\in i^{+};\, \omega \in\Omega_j} \phi^d_{ij} \cdot \mathbb{P}_j(\omega)\cdot V_j(x^\prime, \omega) \\
  \text{s.t.} &(u, x^\prime, y) \in \mathcal{X}_i(x, \omega_i) \\
              & \sum\limits_{d\in D} y_d = 1 \\
              & y_d \in \{0, 1\}, \quad d \in D.
\end{array}
\end{equation}
}
In an equivalent reformulation of~\eqref{eq:dec_dep_formulation_y} we would have constraints $(u, x^\prime) \in \mathcal{X}_i(x, \omega_i)$, but we use the more general $(u, x^\prime, y) \in \mathcal{X}_i(x, \omega_i)$ to allow coupling between $u$ and $y$. We also allow for a more general form $C_i(u, x^\prime, y, \omega_i)$ for reasons that will become clear in Section~\ref{sec:lagrangian_relaxation}.  

Problem formulations with recursions~\eqref{eq:dec_dep_formulation_y} and~\eqref{eq:dec_dep_formulation_y_discrete} bring computational challenges because the products between $y$ (or $y_d$) and $V_j(x^\prime, \omega_j)$ induce non-convexities. In particular, assume the cost-to-go function is convex for fixed $y$. Then, optimizing over $y$ yields the minimum of a set of convex functions, i.e., it yields a non-convex, but piecewise convex cost-to-go function. We revisit this issue via two convex relaxations in Section~\ref{sec:ddu-sddp}, but first we turn to examples.

\subsection{Examples}

To demonstrate the flexibility of our decision-dependent modeling approach, we present a second set of examples.

\begin{example}[The cheese producer]\label{example:cheese}
Consider a small farm that produces cheese. Due to the weather and biological variation, the quantity of cheese produced by the farm each week is the random variable $\omega_q$ [kg].
Each weekend, the farmer can sell cheese at Sunday's market, which has stochastic demand. Any unsold cheese is stored in inventory for the following week. Attending the market costs $c_y$ [\$], and the cheese sells for $c_u$ [\$/kg]. If the farmer attends the market, the demand for cheese is the random variable $\omega_d$ [kg].
The policy graph formulation is given in Figure~\ref{example-fig:cheese}.
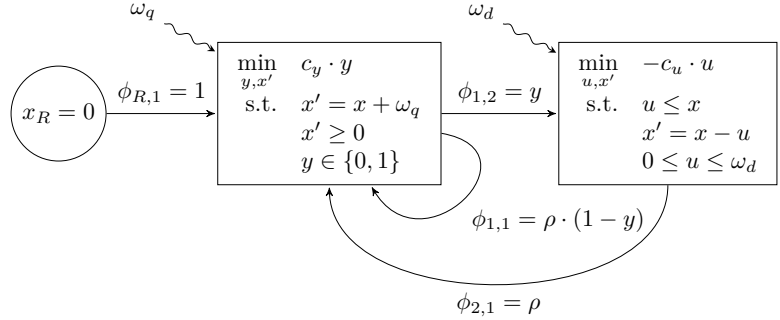
\begin{figure}[ht]
    \centering
\scalebox{0.90}{     
    \begin{tikzpicture}[align=center]
        \node[bellman] (R) at (0, 0) {$x_R = 0$};
        \node[square] (farm)  at (4.0, 0) {
            $\begin{array}{r l}
                \min\limits_{y, x^\prime}& c_y \cdot y \\
              \text{s.t.} & x^\prime = x + \omega_q \\
                          & x^\prime \ge 0 \\
                          & y \in \{0, 1\}
            \end{array}$};
        \node[square] (market)  at (9, 0) {
            $\begin{array}{r l}
                \min\limits_{u, x^\prime}& - c_u \cdot u \\
              \text{s.t.} & u \le x \\
                          & x^\prime = x - u \\
                          & 0 \le u \le \omega_d
            \end{array}$};
        \node[textnode] (w_farm) at (1.25,1.5) {$\omega_q$};
        \node[textnode] (w_market) at (6.25,1.5) {$\omega_d$};
        \path[]
        	(R)	edge 	[] 	node	[left] 	[above] {$\phi_{R,1}=1$}	(farm)
            (farm)	edge 	[] 	node	[left] 	[above] {$\phi_{1,2}= y$}	(market);
        \draw[-stealth, decoration={snake, amplitude = .4mm, segment length = 3mm, post length=1mm},decorate] (w_farm) -- (farm);
        \draw[-stealth, decoration={snake, amplitude = .4mm, segment length = 3mm, post length=1mm},decorate] (w_market) -- 
        (market);
        \draw (market) to [out=-90,in=-90,looseness=1] node {$\phi_{2,1} = \rho$}(farm);
        \draw (farm) to [out=-10,in=-60,looseness=3] node {$\phi_{1,1} = \rho\cdot(1 - y)$}(farm);
    \end{tikzpicture}
}
    \caption{The policy graph for Example~\ref{example:cheese}. This example is a variant of the cyclic newsvendor problem in Examples~\ref{example:two-stage-cyclic} and~\ref{example:markovian}, with random production amounts, along with a binary decision that dictates at which inventory levels we transition to market, i.e., transition to node~2.}
    \label{example-fig:cheese}
\end{figure}
\end{example}

We can form equivalent models for Example~\ref{example:cheese} that do not use decision-dependent transition matrices. For example, we could have added $y$ as a state variable with the constraint $0 \le u \le \omega_d \cdot y$. In general, it is always possible to express a decision-dependent model as a decision-independent model with an expanded state space. We advocate using the decision-dependent policy graph because of the convenience and clarity that it provides the modeler.

\begin{example}[The advertising cheese producer]\label{example:cheese-advertising}
Consider the following variation of Example~\ref{example:cheese}, in which the farmer always attends Sunday's market. During the week, the farmer can advertise for a fixed cost of $c_y$ [\$]. If the farmer advertises, the random demand for cheese is $\omega_d^H$, and 
otherwise the demand is~$\omega_d^L$. 
The policy graph formulation is given in Figure~\ref{example-fig:cheese-advertising}.
\begin{figure}[ht]
    \centering
\scalebox{0.90}{     
    \begin{tikzpicture}[align=center]
        \node[bellman] (R) at (-0.2, 0) {$x_R = 0$};
        \node[square] (farm)  at (3.5, 0) {
            $\begin{array}{r l}
                \min\limits_{y, x^\prime}&  c_y \cdot y \\
              \text{s.t.} & x^\prime = x + \omega_q \\
                          & x^\prime \ge 0 \\
                          & y \in \{0, 1\}
            \end{array}$
        };
        \node[square] (market_l)  at (9, -1.75) {
            $\begin{array}{r l}
                \min \limits_{u, x^\prime}& -c_u \cdot u \\
              \text{s.t.} & u \le x \\
                          & x^\prime = x - u \\
                          & 0 \le u \le \omega_d^L
            \end{array}$};
        \node[square] (market_h)  at (9, 1.75) {
            $\begin{array}{r l}
                \min\limits_{u, x^\prime}& - c_u \cdot u \\
              \text{s.t.} & u \le x \\
                          & x^\prime = x - u \\
                          & 0 \le u \le \omega_d^H
            \end{array}$};
        \node[textnode] (w_farm) at (1.1,1.5) {$\omega_q$};
        \node[textnode] (w_market_l) at (6.25,-0.75) {$\omega_d^L$};
        \node[textnode] (w_market_h) at (6.25,2.75) {$\omega_d^H$};
        \drawsquiggle{w_farm}{farm};
        \drawsquiggle{w_market_l}{market_l};
        \drawsquiggle{w_market_h}{market_h};
        \draw (farm) to [out=-30,in=-180] node [below left] {$1-y$}(market_l);
        \draw (farm) to [out=30,in=-180] node {$y$} (market_h);
        \draw (market_l) to [out=90,in=-5,looseness=0.5] node [below] {$\rho$} (farm);
        \draw (market_h) to [out=-90,in=5,looseness=0.5] node [above] {$\rho$} (farm);
        \draw (R) to [out=0,in=180,looseness=1] node {} (farm);
    \end{tikzpicture}
}
    \caption{The policy graph for Example~\ref{example:cheese-advertising}. This variant of Example~\ref{example:cheese} uses a binary variable $y$ to determine inventory levels at which we should advertise to produce a favorable distribution of demand, $\omega_d^H$ versus $\omega_d^L$, at the market.}
    \label{example-fig:cheese-advertising}
\end{figure}
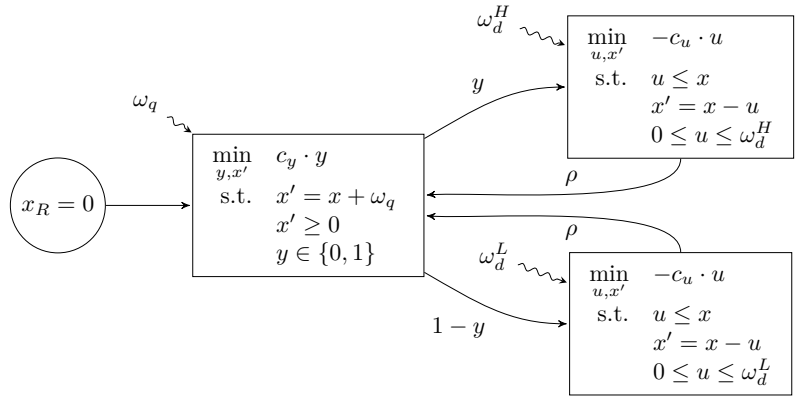
\end{example}

\begin{example}[Advertising]\label{example:advertising}
Consider a Markovian newsvendor problem from Example~\ref{example:markovian}, with the nodes corresponding to \textit{high} and \textit{low} distributions of demand rather than sunny and cloudy conditions, respectively. In each week, the agent can market the product via an advertising budget $y\in \{0, 1\}$. The evolution of the market state forms a Markov chain with a one-step transition matrix that depends on $y$ and includes the discount factor $0 <  \rho < 1$:
$$\Phi(y) = \rho \cdot \begin{bmatrix}
    (0.5+0.3y) & (0.5 - 0.3y) \\ (0.5+0.2y) & (0.5-0.2y)
\end{bmatrix}. $$
\end{example}

\section{Learning}\label{sec:learning}

\subsection{Mathematical model}\label{sec:math_models_learning}

When we must simultaneously learn the reward and transition functions as well as the policy, the requisite statistical learning can occur in a model-based or model-free framework. Here, we consider a setting that requires learning the probability distribution that governs the costs we incur (or, rewards we receive), and the one-step transition matrix. We do so in a structured way, even relative to typical model-based contexts. Once the policy graph model
is formulated, our development adapts ideas from \mbox{Dowson} et al.\ \cite{dowsonPartiallyObservableMultistage2019} for partially observable multi-stage stochastic programs. This section's model also sets up the extension of Section~\ref{sec:math_models_explore_exloit} in which we combine decision-dependent transitions and statistical learning.

\revision{We posit a set of candidate models indexed by $m \in \mathcal{M}$, and we assign a prior belief $b_m > 0$, $m \in \mathcal{M}$, that each is correct, $\sum_{m \in \mathcal{M}} b_m=1$. For computational tractability (see Section~\ref{sec:SDDP_for_dd_learning}'s algorithm), $|\mathcal{M}|$ should be small.} Each model is defined on an identical set of policy graph nodes, denoted by~$\mathcal{N}$.
Model $m \in \mathcal{M}$ has an $|\mathcal{N}| \times |\mathcal{N}|$ one-step transition matrix $\Phi_0^m$, along with $\phi^m_{R,i}$, $i \in \mathcal{N}$, specifying the initial transition probabilities from the root. Model $m$ has a pmf governing the randomness at each node, $i \in \mathcal{N}$, denoted by $\mathbb{P}^m_i(\omega)$, $\omega \in \Omega_i$. We assume that while the one-step transition probabilities can differ, the locations of the nonzero entries are identical across $\Phi_0^m$, $m \in \mathcal{M}$. Moreover, for each model $m \in \mathcal{M}$, while the pmfs can differ, their possible realizations are identical, i.e., they do not depend on~$m$, as the notation $\Omega_i$ indicates. 

To enable statistical learning, we form a larger, augmented policy graph, denoted $\mathcal{G}=(R,\mathcal{N} \times \mathcal{M},\Phi,\mathcal{A})$, which we now describe. We clone nodes across the set of models via $(i,m) \in \mathcal{N} \times \mathcal{M}$. We use a single root node,~$R$. The first row of the $|\mathcal{N} \times \mathcal{M}| + 1$ by $|\mathcal{N} \times \mathcal{M}|$ matrix $\Phi$ is defined by the product $b_m \phi^m_{R,i}$, $m \in \mathcal{M}$ and $i \in \mathcal{N}$, which captures the prior likelihood of each model and the initial transition probabilities within each model. The remaining rows of $\Phi$ are defined by the block-diagonal matrix, $\mbox{diag}\left ( \Phi^m_0 : m \in \mathcal{M} \right )$. The final construct needed is to partition the nodes $\mathcal{N} \times \mathcal{M}$ into \textit{ambiguous subsets}, $A_i=\{(i,m) \, : \, m \in \mathcal{M} \}$ so that $\bigcup_{i \in \mathcal{N}} A_i = \mathcal{N} \times \mathcal{M}$. The agent can observe the current ambiguity set, $A_i$, which is equivalent to simply observing $i$ of the pair $(i,m)$. However, the agent cannot distinguish the models, $m \in \mathcal{M}$, within that ambiguity set. The policy graph $\mathcal{G}$ is augmented by the ambiguity sets, where $\mathcal{A}$ is the partition of the nodes $\mathcal{N} \times \mathcal{M}$ defined by $A_i$, $i \in \mathcal{N}$.

The dynamics in a {\it fully observable} (i.e., a typical) policy graph are as follows. The agent enters node $i$ knowing: (a)~the previous node, $j$; (b)~the incoming physical state, $x$; (c)~the observation, $\omega_i \in \Omega_i$; (d)~the structural constraints, $(u, x') \in \mathcal{X}_i(x,\omega_i)$; (e)~the one-step cost, $C_i(u, x^\prime, \omega_i)$; (f)~the one-step transition probabilities departing $i$, $\phi_{ij}$ for $j \in i^+$ (and subsequent nodes); and (g)~the pmf $\mathbb{P}_j(\omega)$, $\omega \in \Omega_j$ for $j \in i^+$ (and subsequent nodes). 

In our current setting the agent instead enters ambiguity set $A_i$ knowing: (a)~the previous ambiguity set, $A_j$; (b)~the incoming physical state, $x$; (c)~the observation, $\omega_i \in \Omega_i$; (d)~the structural constraints, $(u, x') \in \mathcal{X}_i(x,\omega_i)$; and (e)~the one-step cost, $C_i(u, x^\prime, \omega_i)$. As the notation indicates, $\Omega_i$, $\mathcal{X}_i(x,\omega_i)$, and $C_i(u, x^\prime, \omega_i)$ are each common across all nodes in the ambiguity set $(i,m) \in A_i$. In contrast to the fully observable setting, the agent knows neither (f) nor~(g), but has a belief pmf as to which model is correct. 
Thus the observation $\omega_i \in \Omega_i$ in~(c) now comes from a mixture of $\mathbb{P}_i^m$ with belief weights $b_m$, $m \in \mathcal{M}$. That belief is continually updated via observations. In particular, at the prior stage when in ambiguity set $A_j$, let the agent's belief pmf be $b_m$, $m \in \mathcal{M}$. Upon observing the transition from $A_j$ to $A_i$ and observing $\omega_i \in \Omega_i$ the agent updates the belief from $b$ to $b^\prime$ using Bayes' theorem:
\begin{eqnarray}\label{eq:belief_update}
    b^{\prime}_m &=& \mathbb{P}(m \, | \, \omega_i \ \mbox{and} \ j \rightarrow i ) \nonumber \\ 
    &=&
\frac{\mathbb{P}( \omega_i \ \mbox{and} \ j \rightarrow i \, | \, m) \cdot \mathbb{P}(m)}{\mathbb{P}( \omega_i \ \mbox{and} \ j \rightarrow i)} \nonumber \\ 
& = &
    \frac{
        \mathbb{P}^m_i(\omega_i)\cdot \phi_{ji}^m \cdot b_m
    }{
        \sum\limits_{m \in \mathcal{M}} \mathbb{P}^m_i(\omega_i)\cdot \phi_{ji}^m \cdot b_m
    } .
\end{eqnarray}
We use ``$m$'' to denote the event that $m \in \mathcal{M}$ is the correct model and ``$j \rightarrow i$'' to denote the event that we transitioned from $A_j$ to $A_i$. We let \mbox{$b'=B(b, j \rightarrow i,\omega_i)$} denote the vector-valued update obtained from equation~\eqref{eq:belief_update}.


\revision{Model~\eqref{eqn:full_problem_without_super_pi}'s Bellman reformulation can now be expressed as follows:}
\begin{equation}\label{eqn:full_problem_learn}
\min_{\pi \in \Pi} \sum_{m \in \mathcal{M}} b_m \sum_{i \in R^+}  \phi_{R,i}^m \sum_{\omega \in\Omega_i}\mathbb{P}_i^m(\omega)\cdot V_{i}(x_R,B(b,R \rightarrow i,\omega),\omega),
\end{equation}
where:
\begin{equation}\label{eq:belief_sp_learn}
\begin{array}{l r l}
   &V_{i}(x, b, \omega_i)= \displaystyle \min_{u, x^\prime} & C_i(u, x^\prime, \omega_i) + \mathcal{V}_i(x^\prime, b) \\
   & \mbox{s.t} & (u, x') \in \mathcal{X}_i(x, \omega_i),
\end{array}
\end{equation}
and where:
\begin{equation*}
\mathcal{V}_i(x^\prime, b) = \sum_{m \in\mathcal{M}} b_m \sum_{j \in i^+} \phi_{ij}^m \sum_{\omega \in \Omega_j} \mathbb{P}^m_j (\omega) \cdot V_j(x^\prime, B(b,i \rightarrow j, \omega), \omega) .
\end{equation*}

Thus, we have formulated a policy graph model with ambiguous subsets that captures competing models, $m \in \mathcal{M}$, of our primitives, i.e., competing models of the one-step transition matrix and the randomness at each policy graph node. 
{This formulation incorporates statistical learning into the policy graph framework.}

The simplicity of the Bayesian update~\eqref{eq:belief_update} relative to that in~\cite{dowsonPartiallyObservableMultistage2019} comes from the model structure. If we had a Markov switching model---in which transitions among models $m \in \mathcal{M}$ were possible over time---then we would not have the block-diagonal structure within $\Phi$ discussed above, and the numerator in~\eqref{eq:belief_update} would involve a sum across multiple models. 

\revision{Under model misspecification, the update from Bayes' theorem~\eqref{eq:belief_update} can become overconfident in the wrong model. A strategy to avoid that is tempering, which forms a convex combination of the Bayesian update and the previous belief $b_m$~\cite{bissiri2016general,ibrahim2000power,o1995fractional}. The convex combination's weight specifies a learning rate, which can be made adaptive to account for misspecification or nonstationarity~\cite{grunwald2012safe}.  Our framework could accommodate such an update in case robustness considerations are needed.
}


There is analogous work in the MDP literature involving unknown transition probabilities, hidden context, and competing models \cite{bielecki2023risk,ghatrani2023inverse,hallak2015contextual,steimleMultiModel2021}, and via control of time-staged hidden Markov models, e.g., \cite{elliott2008hidden}, that capture, for example, hidden bear-neutral-bull states of the market in financial applications~\cite{mamon2007hidden}. {These ideas} inspire what we propose here for our class of MDPs, which we can solve using adaptations of SDDP; \revision{see Section~\ref{sec:sddp_whole_section}.}

 


\subsection{Examples}

An example motivates our approach to statistical learning. We defer further examples to \revision{Section~\ref{sec:dd_learning} where we formulate models with decision-dependent learning.}

\begin{example}[Markovian newsvendor with learning]\label{example:learning}
In Example~\ref{example:markovian}, we sketch a newsvendor model in which we have Markov switching between sunny and cloudy states, with a one-step transition matrix,
\begin{equation*}
    \Phi = \begin{bmatrix}
     \phi_{R,s} & \phi_{R,c}  \\
     \phi_{s,s} & \phi_{s,c} \\
     \phi_{c,s} & \phi_{c,c} \\
    \end{bmatrix}. 
\end{equation*}
\revision{We assume all entries of $\Phi$ are positive, $\phi_{s,s}+\phi_{s,c} < 1$, and $\phi_{c,s}+\phi_{c,c} < 1$, so that assumption (A5) holds.} 
Here, we extend the example to the statistical learning setting of Section~\ref{sec:math_models_learning} using Figure~\ref{example-fig:learning}. The left and right submodels clone the underlying policy graph using two candidate models, $\mathcal{M}=\{1,2\}$, with prior belief $b_1>0$ and $b_2>0$, $b_1+b_2=1$. With new node set $\mathcal{N} \times \mathcal{M}=\{(s,1),(c,1),(s,2),(c,2)\}$, the one-step transition matrix is: 
\begin{equation*}
    \Phi = \begin{bmatrix}
     b_1 \phi^1_{R,s} & b_1 \phi^1_{R,c} & b_2 \phi^2_{R,s} & b_2 \phi^2_{R,c}  \\
     \phi^1_{s,s} & \phi^1_{s,c} & 0 & 0 \\
     \phi^1_{c,s} & \phi^1_{c,c} & 0 & 0 \\
      0 & 0 & \phi^2_{s,s} & \phi^2_{s,c}  \\
      0 & 0 & \phi^2_{c,s} & \phi^2_{c,c}  \\
    \end{bmatrix}, 
\end{equation*}
where the superscripts 1 and 2 correspond to the two models. The ambiguity sets are $A_{s}=\{(s,1), (s,2) \}$ and $A_{c}=\{(c,1), (c,2) \}$, so the agent is aware whether it is sunny or cloudy but is unaware whether the correct model is left (1) or right (2). The agent sequentially observes the transitions among the sunny and cloudy nodes and observes realizations of demands at those nodes, which have common support 
across the two models but have different pmfs. At each time step the agent updates the belief pmf, $b$, via equation~\eqref{eq:belief_update}, learning which model is correct.

\begin{figure}[ht]
    \centering
\scalebox{0.90}{     
    \begin{tikzpicture}[align=center]
        \node[bellman] (a) at (2, 0.0) {$R$};
        \node[square, fill=yellow] (s1)  at (0, 1.5) {$(s,1)$};
        \node[square, fill=gray!60] (c1)  at (0, -1.5) {$(c,1)$};
        \node[square, fill=yellow] (s2)  at (4, 1.5) {$(s,2)$};
        \node[square, fill=gray!60] (c2)  at (4, -1.5) {$(c,2)$};
        \node[textnode] (w_s1) at (-0.6,2.4) {$\omega_{s}^1$};
        \node[textnode] (w_c1) at (-0.6,-0.6) {$\omega_{c}^1$};
        \node[textnode] (w_s2) at (3.4,2.4) {$\omega_{s}^2$};
        \node[textnode] (w_c2) at (3.4,-0.6) {$\omega_{c}^2$};
        \path[]
        (a)	edge 	[] 	node	[left] 	[above] {}	(s1)
        (a)	edge 	[] 	node	[left] 	[below] {}	(c1)
        (a)	edge 	[] 	node	[left] 	[above] {}	(s2)
        (a)	edge 	[] 	node	[left] 	[below] {}	(c2)
        ;
        \draw[-stealth, decoration={snake, amplitude = .4mm, segment length = 1.5mm, post length=0.9mm},decorate] (w_s1) -- (s1);
        \draw[-stealth, decoration={snake, amplitude = .4mm, segment length = 1.5mm, post length=0.9mm},decorate] (w_c1) -- (c1);
        \draw[-stealth, decoration={snake, amplitude = .4mm, segment length = 1.5mm, post length=0.9mm},decorate] (w_s2) -- (s2);
        \draw[-stealth, decoration={snake, amplitude = .4mm, segment length = 1.5mm, post length=0.9mm},decorate] (w_c2) -- (c2);
        \draw (c1) to [out=100,in=-100] node {$\phi^1_{c,s}$}(s1);
        \draw (c1) to [out=0,in=-90,looseness=8] node {$\phi^1_{c,c}$}(c1);
        \draw (s1) to [out=0,in=90,looseness=8] node [above right] {$\phi^1_{s,s}$} (s1);
        \draw (s1) to [out=-80,in=80] node {$\phi^1_{s,c}$}(c1);
        \draw (c2) to [out=100,in=-100] node {$\phi^2_{c,s}$}(s2);
        \draw (c2) to [out=0,in=-90,looseness=8] node {$\phi^2_{c,c}$}(c2);
        \draw (s2) to [out=0,in=90,looseness=8] [above right] node {$\phi^2_{s,s}$}(s2);
        \draw (s2) to [out=-80,in=80] node {$\phi^2_{s,c}$}(c2);      
    \end{tikzpicture}
}
    \caption{The policy graph for Example~\ref{example:learning}. The agent is aware of whether the current node is sunny~(yellow) or cloudy~(gray), but is unaware of whether model $m=1$~(left) or \mbox{$m=2$}~(right) is correct. The one-step transition probabilities differ in the left and right halves of the graph, as illustrated by the $\Phi^1$ and $\Phi^2$ labels on arcs. While the supports in the two models are identical, their pmfs differ. The agent observes transitions between sunny and cloudy ambiguity sets, observes demands, and iteratively updates the belief pmf using Bayes' rule~\eqref{eq:belief_update}.}
    \label{example-fig:learning}
\end{figure}
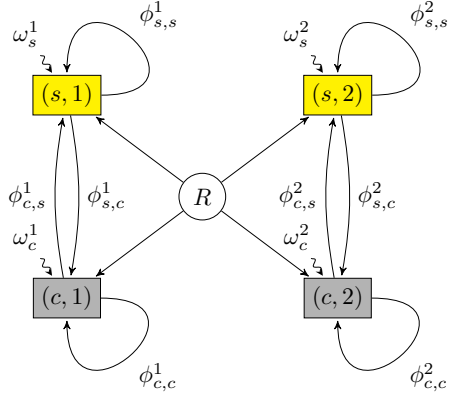
\end{example}

\section{Decision-dependent learning}\label{sec:dd_learning}

\subsection{Mathematical model}\label{sec:math_models_explore_exloit}

We formulate a decision-dependent learning model, combining ideas from Sections~\ref{sec:decision_dep_math_models} and~\ref{sec:math_models_learning}. That is, we allow for action-dependent one-step transition probabilities, which can inform our statistical learning of which probabilistic model is correct: 
\begin{equation}\label{eq:belief_dd_learn}
\min_{\pi \in \Pi} \sum_{m \in \mathcal{M}} b_m \sum_{i \in R^+} \phi_{R,i}^m \sum_{\omega \in\Omega_i}\mathbb{P}_i^m(\omega)\cdot V_{i}(x_R,B_R(b,R \rightarrow i,\omega),\omega)
\end{equation}
where:
\begin{equation}\label{eq:belief_sp_dd_learn}
\begin{array}{l r l}
   &V_{i}(x, b, \omega_i)= \displaystyle \min_{u, x^\prime,y} & C_i(u, x^\prime, y, \omega_i) + \mathcal{V}_i(x^\prime, y, b) \\
   & \mbox{s.t} & (u, x') \in \mathcal{X}_i(x, \omega_i) \\
  && y \in \mathcal{Y},
\end{array}
\end{equation}
and where:
\begin{equation*}
\mathcal{V}_i(x^\prime, y, b) = \sum_{m \in\mathcal{M}} b_m \sum_{j \in i^+} \phi_{ij}^m (y) \sum_{\omega \in \Omega_j} \mathbb{P}^m_j (\omega) \cdot V_j(x^\prime, B(y,b,i \rightarrow j, \omega), \omega) .
\end{equation*}
Here, $B_R(b,R \rightarrow i,\omega)$ has the form of equation~\eqref{eq:belief_update} because there is no $y$-decision at the root node, and $B(y,b,j \rightarrow i, \omega_i)$ is the vector form of:
\begin{eqnarray}\label{eq:belief_update_dec_dep}
    b^{\prime}_m &=& 
    \frac{
        \mathbb{P}^m_i(\omega_i)\cdot \phi_{ji}^m(y) \cdot b_m
    }{
        \sum\limits_{m \in \mathcal{M}} \mathbb{P}^m_i(\omega_i)\cdot \phi_{ji}^m(y) \cdot b_m
    } .
\end{eqnarray}
We update the belief pmf given that we have observed a transition from ambiguity set $A_j$ to $A_i$, observed realization $\omega_i$ in $A_i$, and given that decision $y$ was made in node $j$, affecting the transition probabilities via $\phi_{ji}^m(y)$. 

\subsection{Examples}

We present three examples, which help motivate the formulation above.

\begin{example}[Marketing discovery]
Let $\phi_{ij}$ capture transitions between favorable and less favorable states in a Markov-switching model. Thus, as discussed in Example~\ref{example:advertising}, we can think of $y$ as a marketing decision that may be used to increase the likelihood, via $\phi_{ij}(y)$, of transitioning to a favorable state. Now, however, we may be unsure of the effectiveness of marketing, with it being more effective under model $m=1$, $\phi^1_{ij}(y)$, and less effective under $m=2$,~$\phi^2_{ij}(y)$. Moreover, we may be unsure of {\it how} favorable the demand distribution $\mathbb{P}^1$ is relative to $\mathbb{P}^2$. By investing in potentially costly marketing, we can learn the effectiveness and degree of favorability. An example decision-dependent one-step transition matrix is:
$$\Phi(y) = \rho \cdot \begin{bmatrix}
    0.25 & 0.25 & 0.25 & 0.25 \\
    (0.5+0.3y) & (0.5-0.3y) & 0 & 0 \\
    (0.5+0.2y) & (0.5-0.2y) & 0 & 0 \\
    0 & 0 & (0.5+0.1y) & (0.5 - 0.1y) \\
    0 & 0 & (0.5+0.1y) & (0.5-0.1y)\\
\end{bmatrix},$$
where we have a binary marketing decision $y \in \mathcal{Y} = \{0,1\}$ at each node.    
\end{example}

\begin{example}[Decision-dependent information discovery]\label{example:cheese-information-discovery}
Consider the following variation of Example~\ref{example:cheese}. When starting their business, the farmer has two hypotheses for demand for cheese, represented by the distributions of $\omega^H_d$ and $\omega^L_d$, which are equally likely. Each week, the farmer can pay to attend the market and observe one demand realization. With this information, the farmer updates their belief pmf for whether the demand is from $\omega^H_d$ or $\omega^L_d$. If the farmer does not attend the market, no new information is learned, and the belief pmf is not updated. The policy graph formulation is given in Figure~\ref{example-fig:cheese-information-discovery}.
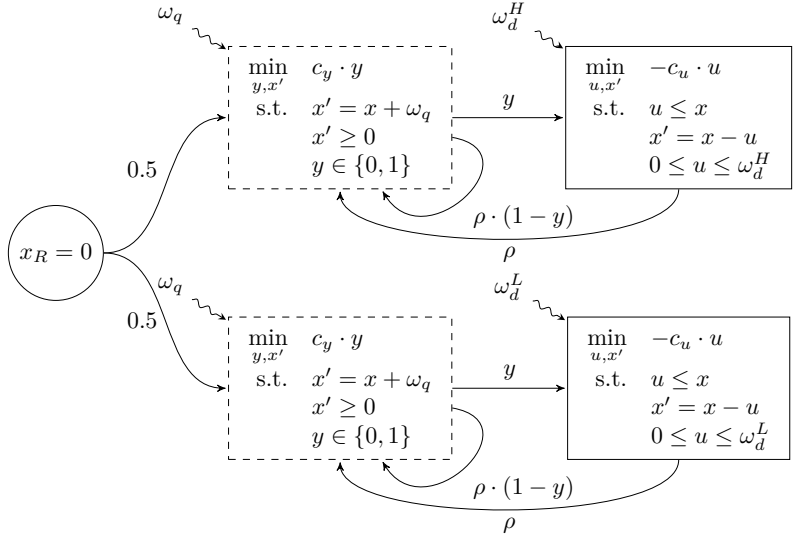
\begin{figure}[ht]
    \centering
\scalebox{0.90}{    
    \begin{tikzpicture}[align=center]
        \node[bellman] (R) at (-0.2, 0) {$x_R = 0$};
        \node[square,dashed] (farm1)  at (3.5, -2) {
            $\begin{array}{r l}
                \min\limits_{y, x^\prime}&  c_y \cdot y \\
              \text{s.t.} & x^\prime = x + \omega_q \\
                          & x^\prime \ge 0 \\
                          & y \in \{0, 1\}
            \end{array}$};
        \node[square] (market1)  at (8, -2) {
            $\begin{array}{r l}
                \min \limits_{u, x^\prime}& - c_u \cdot u \\
              \text{s.t.} & u \le x \\
                          & x^\prime = x - u \\
                          & 0 \le u \le \omega_d^L
            \end{array}$};
        \node[textnode] (w_farm1) at (1.5,-0.5) {$\omega_q$};
        \node[textnode] (w_market1) at (5.5,-0.5) {$\omega_d^L$};
        \drawsquiggle{w_farm1}{farm1};
        \drawsquiggle{w_market1}{market1};
        \draw (farm1) to [out=0,in=-180] node {$y$}(market1);
        \draw (market1) to [out=-90,in=-90,looseness=0.5] node {$\rho$}(farm1);
        \draw (farm1) to [out=-10,in=-60,looseness=2.5] node {$\rho\cdot(1 - y)$}(farm1);
        \node[square,dashed] (farm2)  at (3.5, 2) {
            $\begin{array}{r l}
                \min \limits_{y, x^\prime}& c_y \cdot y \\
              \text{s.t.} & x^\prime = x + \omega_q \\
                          & x^\prime \ge 0 \\
                          & y \in \{0, 1\}
            \end{array}$};
        \node[square] (market2)  at (8, 2) {
            $\begin{array}{r l}
                \min \limits_{u, x^\prime}& -c_u \cdot u \\
              \text{s.t.} & u \le x \\
                          & x^\prime = x - u \\
                          & 0 \le u \le \omega_d^H
            \end{array}$};
        \node[textnode] (w_farm2) at (1.5,3.5) {$\omega_q$};
        \node[textnode] (w_market2) at (5.5,3.5) {$\omega_d^H$};
        \drawsquiggle{w_farm2}{farm2};
        \drawsquiggle{w_market2}{market2};
        \draw (market2) to [out=-90,in=-90,looseness=0.5] node {$\rho$}(farm2);
        \draw (farm2) to [out=-10,in=-60,looseness=2.5] node {$\rho\cdot(1 - y)$}(farm2);
        \draw (farm2) to [out=0,in=-180] node {$y$}(market2);
        \draw (R) to [out=0,in=180,looseness=1] node {$0.5$}(farm2);
        \draw (R) to [out=0,in=180,looseness=1] node [left] {$0.5$}(farm1);
    \end{tikzpicture}
}    
    \caption{The policy graph for Example~\ref{example:cheese-information-discovery}. The two dashed nodes form an ambiguity set, and the two solid nodes form an ambiguity set.}
    \label{example-fig:cheese-information-discovery}
\end{figure}
\end{example}

\begin{example}[Tiger problem]\label{example:tiger}
Consider the following problem from Section 5.1 of~\cite{kaelbling_planning_1998}. An agent is in a room with two exit doors. Behind one is a tiger and the other is escape. In each period, the agent may listen for the tiger or open one of the doors. Opening the door to the tiger costs \$100 (in medical bills). Finding the escape is a reward of \$10. Each turn listening costs \$1.  Because of the muffled sound, if the agent listens for the tiger and it appears to come from a particular door, there is a 15\% chance that it is a false positive and the tiger is behind the other door. The true-positive and false-positive events are independent across periods. Thus, at the per-period listening cost, the agent can improve estimation of the tiger's location by listening in multiple periods.

The problem can be cast as our decision-dependent learning model~\eqref{eq:belief_dd_learn}; see Figure~\ref{example-fig:tiger}. There are no state or control variables in the model, only the decision-dependent $y$. The objective function of the $l$ and $r$ nodes is the constant $1$ (representing the cost of listening), the objective function of the $ll$ and $rr$ nodes is the constant $100$ (representing opening the door to the tiger), and the objective function of the $lr$ and $rl$ nodes is the constant $-10$ (representing opening the door to escape).


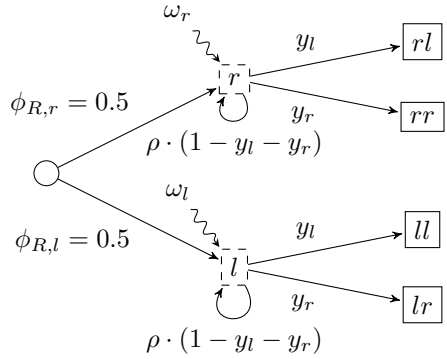
\begin{figure}[!ht]
    \centering   
    \begin{tikzpicture}[align=center,node distance=2cm]
        \node[bellman] (a) at (0, 0) {};
        \node[square,dashed] (l)  at (2, -1) {$l$};
        \node[textnode] (l_w)  at (1.75, -0.25) {$\omega_l$};
        \node[square] (ll)  at (4, -0.5) {$ll$};
        \node[square] (lr)  at (4, -1.5) {$lr$};
        \node[square,dashed] (r)  at (2, 1) {$r$};
        \node[textnode] (r_w)  at (1.75, 1.75) {$\omega_r$};
        \node[square] (rl)  at (4, 1.5) {$rl$};
        \node[square] (rr)  at (4, 0.5) {$rr$};
        \path[]
        	(a)	edge 	[] 	node	[left] 	[below left] {$\phi_{R,l}=0.5$}	(l)
            (a)	edge 	[] 	node	[left] 	[above left] {$\phi_{R,r}=0.5$}	(r)
            (l)	edge 	[] 	node	[left] 	[above left] {$y_l$}	(ll)
            (l)	edge 	[] 	node	[left] 	[below left] {$y_r$}	(lr)
            (r)	edge 	[] 	node	[left] 	[above left] {$y_l$}	(rl)
            (r)	edge 	[] 	node	[left] 	[below left] {$y_r$}	(rr);
        \draw (l) to [out=-60,in=-120,looseness=6] node {$\rho \cdot (1 - y_l - y_r)$}(l);
        \draw (r) to [out=-60,in=-120,looseness=6] node {$\rho \cdot (1 - y_l - y_r)$}(r);
        \draw[-stealth, decoration={snake, amplitude = .6mm, segment length = 2mm, post length=0.9mm},decorate] (l_w) -- (l);
        \draw[-stealth, decoration={snake, amplitude = .6mm, segment length = 2mm, post length=0.9mm},decorate] (r_w) -- (r);
    \end{tikzpicture}
    \caption{The tiger problem formulated as a decision-dependent learning model~\eqref{eq:belief_dd_learn} with $y_l + y_r \le 1, y_l, y_r \in \{0,1\}$. The ambiguity set is $\mathcal{A} = \{\{l, r\}, \{rl\}, \{rr\}, \{ll\}, \{lr\}\}$.}
    \label{example-fig:tiger}
\end{figure}

\end{example}

\section{Stochastic dual dynamic programming}\label{sec:sddp_whole_section}

This section outlines stochastic dual dynamic programming (SDDP), the algorithm that \revision{we use to approximately solve} the models we propose. We begin with the basic algorithm, which \revision{extends nested Benders' decomposition to problems with many stages and has been extensively studied (see, for example, the recent review of \cite{fullner2021stochastic}), and which} can handle the formulation of Section~\ref{sec:policy_graphs_math_models}. We then \revision{provide new extensions of} that algorithm to cover models with decision-dependent transitions and decision-dependent learning. {We study SDDP algorithms because under convexity assumptions they asymptotically  find optimal policies to problems with continuous action and state variables without needing discretization \cite{philpott_convergence_2008}. We say \textit{approximately solve} for two reasons. Even under convexity, we must terminate the sampling-based algorithms finitely, but more importantly, the extensions we propose introduce non-convexities that violate assumptions required for global convergence.}

\subsection{SDDP basics}\label{sec:sddp}

By \revision{assumptions (A1)--(A5) from Section~\ref{sec:assumptions},} the expected cost-to-go function in problem~\eqref{eq:DH} is convex in $x^\prime$ and hence can be approximated from below by the maximum of a set of affine functions known as \textit{cuts}:
$$\E\limits_{j\in i^{+};\, \omega_j} \left[V_j(x^\prime, \omega_j)\right] \ge \max\limits_{k \in \cK} \left\{\alpha_{i,k} + \beta_{i,k}^\top \, x^\prime  \right\},$$
where $\cK=\{0,1,\ldots,K\}$. We assume $\alpha_{i,0}=-M$ and $\beta_{i,0}=0$ to preclude unboundedness of the subproblems we build, where $M$ is sufficiently large.

SDDP iteratively forms cuts that give a polyhedral outer-approximation of the expected cost-to-go function at each node $i \in \mathcal{N}$, exploiting independence of $\omega_i$, $i \in \mathcal{N}$, and the Markovian nature of the one-step transitions. After $K$ iterations of SDDP we approximate~\eqref{eq:DH} as follows:
\begin{equation*}
\begin{array}{r l l}
    \mathbf{SP}_i^K: \quad V^K_i(\bar{x}, \omega_i) =
     \min\limits_{u, x, x^\prime, \theta}&  C_i(u, x^\prime, \omega_i) + \theta \\
   \text{s.t.} & (u, x, x^\prime) \in \mathcal{X}_i(\omega_i) \\
          & \theta \ge \alpha_{i,k} + \beta_{i,k}^{\top} \, x^\prime, &\quad k \in \cK \\
          & x = \bar{x} &\quad [\lambda].
\end{array}
\end{equation*}
The constraints include $(u, x, x^\prime) \in \mathcal{X}_i(\omega)$ \revision{(see assumption~(A4))} because we added constraint $x = \bar{x}$. This \textit{fishing constraint} both simplifies assessing convexity and computing cuts because the incoming state,~$\bar{x}$, appears only in the right-hand side of a linear constraint. Thus, an optimal dual variable $\lambda$ is a subgradient of $V^K_i(\cdot, \omega_i)$ at $\bar{x}$.

Each iteration of SDDP has two phases: (i)~a forward pass, which, by sampling a sequence of nodes and realizations of the random variables, generates a sequence of state variables; and (ii)~a backward pass, which constructs a new cut at each value of the state variables from the forward pass. 

Algorithm~\ref{alg:sddp} gives pseudo-code for SDDP's $K^{th}$ iteration. \revision{If the graph is linear as in Figure~\ref{example-fig:standard_T_stage_MSP} for a standard $T$-stage MSP then Algorithm~\ref{alg:sddp} is equivalent to the original SDDP algorithm of \cite{pereira_multi-stage_1991} with one forward pass per iteration. The extension to general graphs was developed in \cite{dowson_policy_2018}.}

The while loop yields a forward pass, simulating the performance of the policy defined by iteratively solving $\textbf{SP}{}_i^{K-1}(\bar{x},\omega)$ over a sample path defined by nodes $i \in \mathcal{N}$ (sampled from $\Phi$) and realizations $\omega$ (sampled from $\omega_i$'s pmf). The $rand()$ function gives a uniform $(0, 1)$ random variable and accounts for the implicit termination of the \revision{absorbing Markov chain}.
By performing multiple forward passes with a fixed set of cuts for each $\textbf{SP}{}_i^{K-1}$ we can form a \revision{sample mean estimator} of that policy's expected cost.

\revision{In Algorithm~\ref{alg:sddp}'s backward pass,} we iterate through list $\mathcal{S}$, from the last element back to the first, via $reverse(\mathcal{S})$, i.e., reversing the sample path. The backward pass {\it trains the policy}, i.e., refines the set of cuts that approximate the expected cost-to-go function at nodes encountered on the forward pass. In the pseudo-code, we append an iteration index $K$ to $\bar{x}$, the objective function value, and dual variable~$\lambda$ to emphasize the iteration dependence in each cut. 


\begin{algorithm}[!ht]
\scriptsize
    \SetAlgoLined
    \tcc{Forward pass}
    set $\bar{x}=x_R$, $\mathcal{S} = [\;]$, $i = R$\\
    \While{$rand() < \sum{}_{j \in i^+} \phi_{ij} 
    $}{
        sample new $i$ from $i^+$ according to pmf, $\phi_{i, \cdot}/\sum_{j \in i^+} \phi_{ij}$ \\
        sample $\omega$ from $\Omega_i$ according to pmf $\mathbb{P}_i$\\
        solve \textbf{SP}${}_i^{K-1}(\bar{x},\omega)$ and obtain $x^\prime$ \\
        set $\bar{x} = x^\prime$\\
        append $(i, \bar{x})$ to the list $\mathcal{S}$\\
    }
    \tcc{Backward pass}
    \For{$(i, \bar{x}^K)$ in $reverse(\mathcal{S})$}{
        \If {$i^+ = \varnothing$}{
            continue to next {\bf for}-loop iterate of $reverse(\mathcal{S})$\\
        }
        \For{$j\in i^+$, $\omega \in\Omega_j$}{
            solve \textbf{SP}${}_j^{K-1}(\bar{x}^K,\omega)$ \\
            set $V^{K-1}_j(\bar{x}^K, \omega)$ to the optimal objective function value\\
            set $\bar{\lambda}^K_{j,\omega}$ to an optimal dual variable $\lambda$ \\
        }
        set $\beta_{i,K} = \E_{j\in i^+; \, \omega_j} \left[\bar{\lambda}^K_{j,\omega_j}\right]$\\
        set $\alpha_{i,K} = \E_{j\in i^+; \, \omega_j} \left[V^{K-1}_j(\bar{x}^K, \omega_j)\right] - \beta_{i,K}^\top \, \bar{x}^K$ \\
        add the cut $\theta \ge \alpha_{i,K} + \beta_{i,K}^\top \, x^\prime$ to create \textbf{SP}${}_{i}^K$
    }
    \caption{$K^{th}$ iteration of SDDP under recursion~\eqref{eq:DH}.}\label{alg:sddp}
\end{algorithm}

\subsection{SDDP for decision-dependent transitions}\label{sec:ddu-sddp}

Section~\ref{sec:decision_dep_math_models} develops two models that allow $\Phi$ to depend on the agent's actions, yielding non-convex functions $V_i(x,\omega_i)$. Here we derive \revision{novel} \revision{convex relaxations to enable} computation via an SDDP variant. In SDDP's backward pass, we use these relaxations to compute cuts at each node. In SDDP's forward pass we use the {\it non-convex} models, coupled with these policy-defining cuts, \revision{to form} a sample mean upper-bound estimator. Our approximations are relaxations, so we also obtain a lower bound on the model's optimal value, giving a posterior statistical bound on the policy's optimality gap. 

\subsubsection{An adaptive convex relaxation}\label{section:adapative_convex_relax}
We equivalently reformulate recursion~\eqref{eq:dec_dep_formulation_y}, pulling decisions $y$ back one time period, and carrying $y$ into node $i$ in the state. There are two implications, which exploit the fact that in~\eqref{eq:dec_dep_formulation_y} we can restrict attention to polytope~$\mathcal{Y}$'s extreme points (see Section~\ref{sec:decision_dep_math_models}), which we denote $\mbox{ext}(\mathcal{Y}) = \{y_e : \, e \in \mathcal{E}\}$, and assume to be modest in number. The first is that the initial $y$ decision is moved to the root. This is easily handled by making, or enumerating, those decisions at the root. The second is that, rather than a single decision~$y$ at node $i$, we have one for each child $j \in i^+$. The reformulated recursion includes fishing constraints, details $(u, x, x^\prime) \in \mathcal{X}_i(\omega_i)$ by separating the control and transition constraints, expands $V_i$'s arguments to include $y$, and restricts attention to extreme points of $\mathcal{Y}$: \smallskip 
%
{\scalefont{1.0}
\begin{eqnarray}\label{eq:dec_dep_formulation_y2}
    V_i(\bar{x}, \bar{y}, \omega_i) = && \nonumber \\
    \min\limits_{u, x, x^\prime, y, y^\prime}&& C_i( u, x^\prime, \omega_i) + C^\prime_i(y) + \sum\limits_{j\in i^{+}} \phi_{ij}(y) \sum\limits_{\omega \in\Omega_j} \mathbb{P}_j(\omega)\cdot V_j(x^\prime, y^\prime_j,  \omega) \nonumber \\
  \text{s.t.} && u \in U_i(x, \omega_i) \nonumber  \\
&& x^\prime = T_i(u, x, \omega_i) \\
&& y^\prime_j \in \mbox{ext}(\mathcal{Y}),  \ j \in i^+  \nonumber \\
  && x = \bar{x} \nonumber\\
  && y = \bar{y} . \nonumber
\end{eqnarray}
}

\noindent \revision{We simplify (A4) to the case of linear programming, and we revise (A5):}
\begin{enumerate}
    \item[(A4')] \revision{Problem~\eqref{eq:dec_dep_formulation_y2} is a linear program in $(u, x, x^\prime, y, y^\prime)$ when removing the cost-to-go function; $\Phi_0(\cdot)$ satisfies~\eqref{eqn:linear_marketing}; and~$\mathcal{Y}$ is a polytope.}
    \item[(A5')] \revision{For any sequence of $y \in \mbox{ext}(\mathcal{Y})$ the one-step matrices $\Phi_0(\cdot)$ are such that either the policy graph is acyclic or the Bellman operator for problem~\eqref{eq:dec_dep_formulation_y2} is a $\gamma$-contraction.}
\end{enumerate}

Here, \revision{(A4')} relies on the equivalence of~\eqref{eq:dec_dep_formulation_y2} under $y^\prime_j \in \mathcal{Y}$ without the cost-to-go function.
We derive a convex approximation of $V_i(\bar{x}, \bar{y}, \omega_i)$, and compute cuts for SDDP using that approximation. We inductively assume:
\begin{equation}\label{eqn:pg_pwl_approx}
\sum\limits_{\omega \in\Omega_j} \mathbb{P}_j(\omega)\cdot V_j(x^\prime, y^\prime_j,  \omega) \ge 
{\dst \max_{k \in \cK}} \left [ \alpha_{j,k} + \beta^\top_{j,k} \, x^\prime +  \gamma^\top_{j,k} \, y_j^\prime  \right ],
\end{equation}
for $j \in i^+$ where we assume $\cK = \{0,1,2,\ldots,K\}$ after $K$ iterations of SDDP, and we suppress a $j$ index on $\cK$. 
\revision{Using compactness from~(A3) and~(A4'), expression~\eqref{eqn:pg_pwl_approx} holds for $K=0$ with $\beta_{j,0}=\gamma_{j,0}=0$ and with $\alpha_{j,0}=-M$ for $M$ sufficiently large.} 
We show that the form of~\eqref{eqn:pg_pwl_approx} persists as we compute cuts, reversing the sample path from a forward pass, as in Algorithm~\ref{alg:sddp}. 
Our convex lower-bounding approximation is as follows:\smallskip 

\noindent 
{\scalefont{1.00} $V_i(\bar{x}, \bar{y}, \omega_i) \ge V_i^K(\bar{x}, \bar{y}, \omega_i) =$}

\vspace*{-0.2in}

{\scalefont{1.00}
\begin{eqnarray}
{\mathbf{SP}}^K_i:
    \min\limits_{u, x, x^\prime, y, y^\prime,\theta}&& C_i( u, x^\prime, \omega_i)  + C_i^\prime(y)  + \sum\limits_{j\in i^{+}}  \theta_j  \nonumber \\
   \text{s.t.} && u \in U_i(x, \omega_i)  \nonumber \\
                 && x^\prime = T_i(u, x, \omega_i) \label{equation:ddu_first_model_alg} \\
                 && \theta_j  \ge \phi_{ij}(y)\left ( \alpha_{j,k} + \beta^\top_{j,k} \, x^\prime +  \gamma^\top_{j,k} \, y_j^\prime  \right ) , \ \  j \in i^+, k \in \cK  \nonumber \\
&& y^\prime_j \in \mbox{ext}(\mathcal{Y}),  \ j \in i^+  \nonumber \\ 
  && x = \bar{x} \nonumber \\
  && y = \bar{y} \nonumber 
\end{eqnarray}
}
\vspace*{-0.3in}
{\scalefont{0.90}
\begin{eqnarray}
  \ge \ubar{V}^K_i(\bar{x},\bar{y},\omega_i) = \ \nonumber \\  
  \underline{\mathbf{SP}}^K_i:
   \dst  \min\limits_{u, x, x^\prime, y, y^\prime,\theta}&& \dst  C_i \left (  \revision{u} , \, \sum_{e \in \mathcal{E}} x^\prime_e, \, \omega_i \right )  +  C_i^\prime(y) + \sum\limits_{j\in i^{+}}  \theta_j  \nonumber \\
   \text{s.t.} &&  \dst \revision{u} \in U_i(x, \omega_i)  \nonumber \\
                 && \dst \sum_{e \in \mathcal{E}} x^\prime_e = T_i \left ( \revision{u}, \, x, \omega_i \right ) \label{equation:ddu_second_model_alg} \\
                 && \theta_j  - \dst \sum_{e \in \mathcal{E}} \phi_{ij}(y_e) \left ( \beta^\top_{j,k} \, x^\prime_e +  \gamma^\top_{j,k} \, \revision{y_{j,e}^\prime} \right ) \ge \phi_{ij}(y) \cdot {\alpha}_{j,k}  , \ \  j \in i^+, k \in \cK,      \nonumber \\
&& \revision{\sum_{e \in \mathcal{E}} y^\prime_{j,e}} \in \mathcal{Y},  \ j \in i^+  \nonumber \\
  && x = \bar{x}  \ \ \ \ [\lambda_x ] \nonumber \\
  && y = \bar{y}  \ \ \ \ [\lambda_y ] . \nonumber 
\end{eqnarray}
}

The first inequality holds by~\eqref{eqn:pg_pwl_approx}. Decision $\bar{y}$ computed in a forward pass of SDDP is an extreme point of $\mathcal{Y}$, say $\bar{y}=y_{\bar{e}}$. The inequality 
between~\eqref{equation:ddu_first_model_alg} and~\eqref{equation:ddu_second_model_alg} holds because the restriction \revision{$x^\prime_e=0$ and $y^\prime_{j,e}=0$} for $e \in \mathcal{E} \setminus \{ \bar{e} \}$ in the latter yields the former, modulo the relaxation from \revision{constraints involving $\mbox{ext}(\mathcal{Y})$ to those with 
$\mathcal{Y}$}. Under \revision{(A4')},
model~\eqref{equation:ddu_second_model_alg} is a linear program, parameterized in its right-hand side by $\bar{x}$ and~$\bar{y}$. Thus, $\ubar{V}^K_i(\cdot,\cdot,\omega_i)$ is convex.  

In extending Algorithm~\ref{alg:sddp}'s logic, assume that we are computing the $K$-th cut {\it from} node~$j$ with incoming state $(\bar{x}^K,\bar{y}^K)$. Then from subproblem~\eqref{equation:ddu_second_model_alg}, with $i$ replaced by $j$, we obtain $\bar{\lambda}_{x,j,\omega}^K$ and $\bar{\lambda}_{y,j,\omega}^K$ and compute cuts as follows:
\begin{subequations}\label{eqn:cut_coefficients_ddu_y}
\begin{eqnarray}
\beta_{j,K} & = &  \mathbb{E}_{\omega_j} \left[\bar{\lambda}_{x,j,\omega_j}^K\right] \\
\gamma_{j,K} & = &  \mathbb{E}_{\omega_j} \left[\bar{\lambda}_{y,j,\omega_j}^K\right] \\
{\alpha}_{j,K} & = &  \mathbb{E}_{\omega_j} \left[\ubar{V}^K_j(\bar{x}^K,\bar{y}^K,\omega_j)\right] - \beta^\top_{j,K} \, \bar{x}^K - \gamma^\top_{j,K} \, \bar{y}^K.  
\end{eqnarray}
\end{subequations}
The cuts are an outer linearization of $\mathbb{E}_{\omega_j} \left[\ubar{V}^{K-1}_j(x,y,\omega_j)\right]$.
We summarize our derivation in the following result.

\begin{theorem}\label{theorem:ddu}
Let (A1), (A2), \revision{(A3), (A4'), and (A5')} hold,
assume $y$ is an extreme point of polytope $\mathcal{Y}$, and assume inequality~\eqref{eqn:pg_pwl_approx} holds. Then $$\ubar{V}^K_i(x,y,\omega_i) \le V^K_i(x,y,\omega_i) \le V_i(x,y,\omega_i),$$ where these are the respective optimal values of~\eqref{equation:ddu_second_model_alg}, \eqref{equation:ddu_first_model_alg}, and \eqref{eq:dec_dep_formulation_y2}. We have
$$\ubar{V}^{K-1}_i(x,y,\omega_i) \le \ubar{V}^K_i(x,y,\omega_i) \ \mbox{and} \ 
{V}^{K-1}_i(x,y,\omega_i) \le {V}^K_i(x,y,\omega_i).$$
Moreover, $\ubar{V}^K_i(\cdot,\cdot,\omega_i)$ is convex, and 
$$
\mathbb{E}_{\omega_i} \ubar{V}^K_i(x,y,\omega_i) \ge \max_{k \in \cK} \left [ \alpha_{i,k} + \beta_{i,k}^\top \, x + \gamma_{i,k}^\top \, y \right ],
$$
where $\alpha_{i,k}, \beta_{i,k}, \gamma_{i,k}$ are computed by equation~\eqref{eqn:cut_coefficients_ddu_y} for $i \in \mathcal{N}$, $k \in \cK$. 
\end{theorem}

Algorithm~\ref{alg:ddu_sddp_adaptive} gives the $K$-th iteration of our SDDP variant.  We solve non-convex subproblems $\mathbf{SP}^{K-1}_i$ in forward passes, and solve $\underline{\mathbf{SP}}^{K-1}_i$ in backward passes. In recursion~\eqref{eq:dec_dep_formulation_y}, we can equivalently replace $\mathcal{Y}$ with $\mbox{ext}(\mathcal{Y})$, but not in~\eqref{equation:ddu_first_model_alg} and~\eqref{equation:ddu_second_model_alg}. Solving $\mathbf{SP}^{K-1}_i$ requires binary variables to handle $y^\prime_j \in \mbox{ext}(\mathcal{Y})$ and linearizations for bilinear terms involving binaries. Subproblem $\underline{\mathbf{SP}}^{K-1}_i$ is a linear program. Coupled with~\eqref{eqn:pg_pwl_approx} holding \revision{when $K=0$}, Theorem~\ref{theorem:ddu} shows that Algorithm~\ref{alg:ddu_sddp_adaptive}'s cuts form an outer approximation at each node. 
In general, we obtain a suboptimal policy because of the problem's non-convexity, but we can assess an optimality gap using bounds as discussed above. 



\begin{algorithm}[!ht]
\scriptsize
    \SetAlgoLined
    \tcc{Forward pass}
    set $\bar{x}=x_R$, $\mathcal{S} = [\;]$, $i = R$, $\bar{y}=0$ and $\bar{y}^\prime \in \mathcal{Y}$ \\
    \While{$rand() < \sum{}_{j \in i^+} \phi_{ij}(\bar{y}) 
    $}{
        sample new $i$ from $i^+$ according to pmf, $\frac{\phi_{i, \cdot}(\bar{y})}{\sum_{j \in i^+} \phi_{ij}(\bar{y})}$ \\
        sample $\omega$ from $\Omega_i$ according to pmf $\mathbb{P}_i$ \\
        solve \textbf{SP}${}_i^{K-1}(\bar{x},\bar{y}^\prime,\omega)$ and obtain $x^\prime$ and $y^\prime=[y_j^\prime]_{j \in i^+}$ \\
        append $(i, x^\prime, y^\prime)$ to the list $\mathcal{S}$\\
        set $\bar{x} = x^\prime$, $\bar{y} = \bar{y}^\prime$, $\bar{y}^\prime = y^\prime$\\
    }
    \tcc{Backward pass}
    \For{$(i,\bar{x}^K,\bar{y}^K)$ in $reverse(\mathcal{S})$}{
        \If {$i^+ = \varnothing$}{
            continue to next {\bf for}-loop iterate of $reverse(\mathcal{S})$ \\
        }
        \For{$j\in i^+$}{
        \For{$\omega \in\Omega_j$}{
            solve $\underline{\mathbf{SP}}_j^{K-1}(\bar{x}^K,\bar{y}^K_j,\omega)$ \\
            set $\ubar{V}^{K-1}_j(\bar{x}^K,\bar{y}^K_j,\omega)$ to the optimal objective function value\\
            {{set $\bar{\lambda}^K_{x,j,\omega}$ and $\bar{\lambda}^K_{y,j,\omega}$ to optimal dual variables $\lambda_x$ and $\lambda_y$}} \\
        }
        {{ 
        compute cut coefficients $\alpha_{j,K}, \beta_{j,K},\gamma_{j,K}$ using equation~\eqref{eqn:cut_coefficients_ddu_y}}} \\
        {add the $K^{th}$ cut to \textbf{SP}${}_{i}^{K}$
        and to $\underline{\mathbf{SP}}_{i}^{K}$
        using $\alpha_{j,K}, \beta_{j,K},\gamma_{j,K}$}
        }
    }
    \caption{$K^{th}$ iteration of SDDP under recursion~\eqref{eq:dec_dep_formulation_y}.}\label{alg:ddu_sddp_adaptive}
\end{algorithm}

\subsubsection{Lagrangian relaxation}\label{sec:lagrangian_relaxation}

We succinctly repeat the analysis of Section~\ref{section:adapative_convex_relax} but starting from recursion~\eqref{eq:dec_dep_formulation_y_discrete}, seeking a convex relaxation for $V_i(\bar{x}, \omega_i)$. We inductively assume for $j \in i^+$ and $\omega \in \Omega_j$ that
\begin{equation}\label{eqn:pwl_approx_y_discrete}
V_j(x^\prime, \omega) \ge \max_{k \in \cK} \left [ \alpha_{j,\omega,k} + \beta_{j,\omega,k}^\top \,  x^\prime \right ],
\end{equation}
which holds at \revision{$K=0$ with $\beta_{j,\omega,0}=0$ and $\alpha_{j,\omega,0}=-M$ for $M$ sufficiently large.} Applying~\eqref{eqn:pwl_approx_y_discrete} in recursion~\eqref{eq:dec_dep_formulation_y_discrete}, and adding a fishing constraint $x=\bar{x}$ yields the following analog of subproblem~\eqref{equation:ddu_first_model_alg}: 
\begin{equation}\label{eq:ddu_primal_problem}
\allowdisplaybreaks
\begin{array}{r l}
    \mathbf{SP}^K_i: V_i^K(\bar{x},&\omega_i) = \\
    \min\limits_{u, x, x^\prime, y, \theta, \Theta}& C_i(u, x^\prime, y, \omega_i) + \Theta \\
  \text{s.t.} &(u, x, x^\prime, y) \in \mathcal{X}_i(\omega_i) \\
              & \sum\limits_{d\in D} y_d = 1 \\
              & y_d \in \{0, 1\}, \quad d \in D \\              
              & \Theta \ge \sum\limits_{d \in D} y_d \sum\limits_{j\in i^{+};\ \omega\in\Omega_j} \phi^d_{ij} \cdot \mathbb{P}_j(\omega)\cdot \theta_{j,\omega}\\
              & \theta_{j,\omega} \ge \alpha_{j,\omega,k} + \beta_{j,\omega,k}^\top \, x^\prime,\;  j\in i^+, \omega\in\Omega_j, k\in \cK \\
    & x = \bar{x} \quad [\lambda]. \\
\end{array}
\end{equation}

The constraint for $\Theta$ is nonlinear because binary variable $y_d$ multiplies continuous variable $\theta_{j,\omega}$, but these can again be equivalently linearized. If $C_i$ and $\mathcal{X}_i$ permit a linear formulation, then \eqref{eq:ddu_primal_problem} is a mixed-integer linear program, and $V_i^K(\cdot, \omega_i)$ is non-convex. Because $\mathbf{SP}_i^K$ includes \revision{integer variables, we do} not have immediate access to a subgradient $\lambda$. We could relax integrality when solving $\mathbf{SP}_k^{K-1}$ on the backward pass. Importantly, we would enforce integrality when solving $\mathbf{SP}_k^{K-1}$ on the forward pass. Instead, we take the Lagrangian dual of the $x = \bar{x}$ to obtain the following analog of~\eqref{equation:ddu_second_model_alg}:
\begin{equation}\label{eq:ddu_dual_problem}
\begin{array}{r l}
    \underline{\mathbf{SP}}^K_i: \ubar{V}_i^K(\bar{x},&\omega_i) = \\
    \max\limits_{\lambda} \min\limits_{u, x, x^\prime,  y, \theta, \Theta}& C_i(u, x^\prime, y, \omega_i) + \Theta - \lambda^\top (x - \bar{x}) \\
  \text{s.t.} &(u, x, x^\prime, y) \in \mathcal{X}_i(\omega_i) \\
              & \sum\limits_{d\in D} y_d = 1 \\
              & y_d \in \{0, 1\}, \quad d \in D \\
              & \Theta \ge \sum\limits_{d \in D} y_d \sum\limits_{j\in i^{+};\ \omega\in\Omega_j} \phi^d_{ij} \cdot \mathbb{P}_j(\omega)\cdot \theta_{j,\omega}\\
              & \theta_{j,\omega} \ge \alpha_{j,\omega,k} + \beta_{j,\omega,k}^\top \, x^\prime, \; j\in i^+, \omega\in\Omega_j, k\in \cK.\\
\end{array}
\end{equation}

We do not state the algorithm associated with~\eqref{eq:ddu_primal_problem} 
and~\eqref{eq:ddu_dual_problem} because of its similarity to Algorithm~\ref{alg:ddu_sddp_adaptive}, but we make some comments. With notation parallel to that of Algorithms~\ref{alg:sddp} and~\ref{alg:ddu_sddp_adaptive} the cuts we compute have form: 
$$\theta_{j,\omega} \ge \ubar{V}_j^K(\bar{x}^K,\omega) + \left(\bar{\lambda}^K_{j,\omega}\right)^\top \left(x^\prime - \bar{x}^K\right)$$
so that:
\begin{subequations}\label{eqn:cut_coefficients_ddu_y_discrete}
\begin{eqnarray}
\beta_{j,\omega,K} & =& \bar{\lambda}^K_{j,\omega} \\
\alpha_{j,\omega,K} & =& \ubar{V}_j^K(\bar{x}^K,\omega) - \left ( \bar{\lambda}^K_{j,\omega} \right )^\top \bar{x}^K.
\end{eqnarray}
\end{subequations}
We solve the non-convex problem \eqref{eq:ddu_primal_problem} on the forward pass, and solve the convex relaxation \eqref{eq:ddu_dual_problem} on the backward pass to compute cuts. 

Solving the convex relaxation \eqref{eq:ddu_dual_problem} can be expensive because it requires solving a set of mixed-integer programs at each node for the Lagrangian dual, e.g., using a cutting-plane or subgradient method. That said, we can compute valid but possibly weaker cuts with a few iterations of those methods.
We summarize our convex relaxation in the following result. 
\begin{theorem}\label{theorem:ddu_discrete}
Let (A1), (A2), \revision{(A3), and (A5')} hold, and assume inequality~\eqref{eqn:pwl_approx_y_discrete} holds. Then $$\ubar{V}^K_i(x,\omega_i) \le V^K_i(x,\omega_i) \le V_i(x,\omega_i),$$ where these are the respective optimal values of~\eqref{eq:ddu_dual_problem}, ~\eqref{eq:ddu_primal_problem}, and~\eqref{eq:dec_dep_formulation_y_discrete}. 
We have
$$
\ubar{V}^{K-1}_i(x,\omega_i) \le \ubar{V}^K_i(x,\omega_i) \ \mbox{and} \ 
{V}^{K-1}_i(x,\omega_i) \le {V}^K_i(x,\omega_i).
$$
Moreover, $\ubar{V}^K_i(\cdot,\omega)$ is convex, and 
$$
\ubar{V}^K_i(x,\omega) \ge \max_{k \in \cK} \left [ \alpha_{i,\omega,k} + \beta_{i,\omega,k}^\top \, x  \right ],
$$
where $\alpha_{i,\omega,k}, \beta_{i,\omega,k}$ are computed by equation~\eqref{eqn:cut_coefficients_ddu_y_discrete} for $i \in \mathcal{N}$, $k \in \cK$.  
\end{theorem}

In \revision{(A5')} we use $\Phi(y)=\sum_{d \in D} y_d \Phi^d$, \revision{and~(A4)} is not required because the Lagrangian dual yields a convex relaxation even though the inner minimization in~\eqref{eq:ddu_dual_problem} is non-convex. For this reason, the method could handle additional discreteness or non-convexity in $(u, x, x^\prime, y) \in \mathcal{X}_i(\omega_i)$. 

\subsection{SDDP for decision-dependent learning}\label{sec:SDDP_for_dd_learning}

We parallel Section~\ref{section:adapative_convex_relax}'s development, although we could use Section~\ref{sec:lagrangian_relaxation}. We reformulate~\eqref{eq:belief_sp_dd_learn} to give the analog of~\eqref{eq:dec_dep_formulation_y2}, accounting for the belief pmf:
\begin{eqnarray}\label{eq:dd_learn_formulation_y2}
    V_i(\bar{x}, \bar{y}, b, \omega_i) = && \nonumber \\
    \min\limits_{u, x, x^\prime, y, y^\prime}&& C_i( u, x^\prime, \omega_i) + C^\prime_i(y) + \sum_{\substack{m \in\mathcal{M} \\ j \in i^+ \\ \omega \in \Omega_j}} b_m \cdot \phi_{ij}^m (y) \cdot \mathbb{P}^m_j (\omega) \cdot V_j(x^\prime, y^\prime_j,b^\prime,\omega) \nonumber \\
  \text{s.t.} && u \in U_i(x, \omega_i) \nonumber  \\
&& x^\prime = T_i(u, x, \omega_i) \\
&& y^\prime_j \in \mbox{ext}(\mathcal{Y}),  \ j \in i^+  \nonumber \\
  && x = \bar{x} \nonumber\\
  && y = \bar{y} , \nonumber
\end{eqnarray}
where $b^\prime=B(y,b,i \rightarrow j, \omega)$.  We adapt our assumptions, accounting for both decision-dependence and learning: 
\begin{enumerate}
    \item[(A4'')] \revision{Problem~\eqref{eq:dd_learn_formulation_y2} is a linear program in $(u, x, x^\prime, y, y^\prime)$ when removing the cost-to-go function; $\Phi^m_0(\cdot)$ satisfies~\eqref{eqn:linear_marketing} for all $m \in \mathcal{M}$; and~$\mathcal{Y}$ is a polytope.}
    \item[(A5'')] \revision{For any sequence of $y \in \mbox{ext}(\mathcal{Y})$ and for all $m \in \mathcal{M}$ the one-step matrices $\Phi_0^m(\cdot)$ are such that either the policy graph is acyclic or the Bellman operator for problem~\eqref{eq:dd_learn_formulation_y2} is a $\gamma$-contraction.}
\end{enumerate}

We assume:
\begin{equation}\label{eqn:pg_pwl_dd_learn_approx}
\sum\limits_{\omega \in\Omega_j} \mathbb{P}^m_j(\omega)\cdot V_j(x^\prime, y^\prime_j, b^\prime, \omega) \ge 
{\dst \max_{k \in \cK} \left [ \alpha_{j,m,k} + \beta^\top_{j,m,k} \, x^\prime +  \gamma^\top_{j,m,k} \, y_j^\prime  \right ].}
\end{equation}
%
\noindent Then we can interpolate cuts for different belief states as in~\cite{dowsonPartiallyObservableMultistage2019} to form:
{\scalefont{0.95}
\begin{equation*}
\begin{array}{r l l}
{V}^K_i(\bar{x}, \bar{y}, b, \omega_i)= \\
   \min\limits_{u, x, x^\prime, y, y^\prime, \theta} \max\limits_{\eta \ge 0}& C_i(u, x^\prime, \omega_i) + C_i^\prime(y) + \sum\limits_{k \in \cK} \eta_k \theta_k \\
     \text{s.t.} & u \in U_i(x, \omega_i) \nonumber  \\
& x^\prime = T_i(u, x, \omega_i) \\
           & \theta_{k} \ge \sum\limits_{\substack{m \in \mathcal{M} \\ j \in i^+}} b_{m,k} \cdot \phi_{ij}^m(y) \left ( \alpha_{j,m,k} + \beta_{j,m,k}^\top \, x' + \gamma_{j,m,k}^\top \, y^\prime_j \right ), 
           \ \ k \in \cK \\
        & \sum\limits_{k \in \cK} \eta_{k} b_k = b,  \ \ \ \  [\mu]\\
        & \sum\limits_{k \in \cK} \eta_{k} = 1, \ \ \ \, \ \ \  [\Theta]\\
& y^\prime_j \in \mbox{ext}(\mathcal{Y}),  \ j \in i^+  \nonumber \\
        & x = \bar{x} \ \ \ \  [\lambda_x] \nonumber  \\
        & y = \bar{y} \ \ \ \  [\lambda_y]. \nonumber
\end{array}
\end{equation*}
}

\clearpage

Taking the dual of the inner maximization, we obtain the analog of~\eqref{equation:ddu_first_model_alg}:
{\scalefont{0.95}
\begin{eqnarray}\label{eq:sp_dd_learn}
\allowdisplaybreaks
{\mathbf{SP}}^K_i: \hspace*{0.6in} \nonumber \\
{V}^K_i(\bar{x}, \bar{y}, b, \omega_i)=  \nonumber \\
   \min\limits_{u, x, x^\prime, y, y^\prime, \theta, \mu, \Theta} && C_i(u, x^\prime, \omega_i) + C_i^\prime(y) + b^\top \mu + \Theta  \nonumber \\
     \text{s.t.} && u \in U_i(x, \omega_i) \nonumber  \\
&& x^\prime = T_i(u, x, \omega_i) \nonumber \\
&&   b^\top_k \mu + \Theta \ge \theta_k, \ \ k \in \cK   \\
           && \theta_{k} \ge \sum\limits_{\substack{m \in \mathcal{M} \\ j \in i^+}} b_{m,k} \cdot \phi_{ij}^m(y) \left ( \alpha_{j,m,k} + \beta_{j,m,k}^\top \, x' + \gamma_{j,m,k}^\top \, y^\prime_j \right ), 
           \ \ k \in \cK \nonumber \\
           && y^\prime_j \in \mbox{ext}(\mathcal{Y}),  \ j \in i^+  \nonumber \\
        && x = \bar{x} \ \ \ \  [\lambda_x] \nonumber  \\
        && y = \bar{y} \ \ \ \  [\lambda_y]. \nonumber
\end{eqnarray}
}

\revision{Model~\eqref{eq:sp_dd_learn} is non-convex} but can be solved as a mixed-integer linear program on forward passes of SDDP. Relaxing in the same way as in Section~\ref{section:adapative_convex_relax} yields the analog of~\eqref{equation:ddu_second_model_alg}:

\vspace*{-0.2in}

{\scalefont{0.85}
\allowdisplaybreaks
\begin{eqnarray}\label{eq:sp_underbar_dd_learn}
{\underline{\mathbf{SP}}}^K_i: \hspace*{0.6in} \nonumber \\
\ubar{V}^K_i(\bar{x}, \bar{y}, b, \omega_i)=  \nonumber  \\
   \min\limits_{u, x, x^\prime, y, y^\prime, \theta, \theta^\prime, \mu, \Theta} && C_i(u, x^\prime, \omega_i) + C_i^\prime(y) + b^\top \mu + \Theta \nonumber \\
     \text{s.t.} && u \in U_i(x, \omega_i) \nonumber  \\
&& x^\prime = T_i(u, x, \omega_i) \nonumber  \\
&&   b^\top_k \mu + \Theta \ge \theta_k, \ \ k \in \cK   \\
&& \theta_{k} - \sum\limits_{\substack{e \in \mathcal{E} \\ m \in \mathcal{M} \\ j \in i^+}} b_{m,k} \cdot \phi_{ij}^m(y_e) \cdot  \theta^\prime_{j,m,e,k} \ge \sum\limits_{\substack{m \in \mathcal{M} \\ j \in i^+}} b_{m,k} \cdot \phi_{ij}^m(y) \cdot   \alpha_{j,m,k}, 
            k \in \cK \nonumber  \\
&& \theta^\prime_{j,m,e,k} \ge \beta_{j,m,k}^\top \, x'_e + \gamma_{j,m,k}^\top \, \revision{y^\prime_{j,e}}, \ \ j \in i^+, m \in \mathcal{M}, e \in \mathcal{E}, k \in \cK  \nonumber \\
        &&  x^\prime = \sum\limits_{e \in \mathcal{E}} x^\prime_e \nonumber \\
        && \revision{\sum\limits_{e \in \mathcal{E}} y^\prime_{j,e}} \in \mathcal{Y},  \ j \in i^+  \nonumber \\
        && x = \bar{x} \ \ \ \  [\lambda_x] \nonumber  \\
        && y = \bar{y} \ \ \ \  [\lambda_y]. \nonumber
\end{eqnarray}
}
Under (A4''), model~\eqref{eq:sp_underbar_dd_learn} is a linear program, which is parameterized in its right-hand side by $(\bar{x},\bar{y})$ and its objective function by $b$. Thus, $\ubar{V}^K_i(\bar{x}, \bar{y}, b, \omega_i)$ is convex in $(\bar{x},\bar{y})$ for fixed $b$ and concave in $b$ for fixed $(\bar{x},\bar{y})$.

With $\mathbb{E}_{\omega^m_j} [ \cdot ]$ denoting $\sum_{\omega \in \Omega_j} \mathbb{P}^m_j(\omega) [ \cdot ]$ we compute cut coefficients:
\begin{subequations}\label{eqn:cut_coefficients_ddu_y_learn}
\begin{eqnarray}
\beta_{j,m,K} & = &  \dst {\mathbb{E}}_{\omega^m_j} \left[\bar{\lambda}_{x,j,\omega_j^m}^K\right] \\
\gamma_{j,m,K} & = &  \dst {\mathbb{E}}_{\omega^m_j} \left[\bar{\lambda}_{y,j,\omega_j^m}^K\right] \\
{\alpha}_{j,m,K} & = &  \dst {\mathbb{E}}_{\omega^m_j} \left[\ubar{V}^K_j(\bar{x}^K,\bar{y}^K, b^K, \omega_j^m)\right] - \beta^\top_{j,m,K} \, \bar{x}^K - \gamma^\top_{j,m,K} \, \bar{y}^K.  
\end{eqnarray}
\end{subequations}

\noindent The cuts are an outer linearization of 
$$
\sum_{\omega \in \Omega_j} \mathbb{P}^m_j (\omega) \cdot \ubar{V}^K_j(x^\prime, y^\prime, b^\prime, \omega).
$$ 
The above development yields the following result.
\begin{theorem}\label{theorem:ddu-belief}
Let (A1), (A2), \revision{(A3), (A4''), and (A5'')} hold,
assume $y$ is an extreme point of polytope $\mathcal{Y}$, and assume inequality~\eqref{eqn:pg_pwl_dd_learn_approx} holds. Then $$\ubar{V}^K_i(x,y,b,\omega_i) \le V^K_i(x,y,b,\omega_i) \le {V_i(x,y,b,\omega_i)},$$ where these are the respective optimal values of~\eqref{eq:sp_underbar_dd_learn}, \eqref{eq:sp_dd_learn}, and \eqref{eq:dd_learn_formulation_y2}. 
We have
$$\ubar{V}^{K-1}_i(x,y,b,\omega_i) \le \ubar{V}^K_i(x,y,b,\omega_i) \ \mbox{and} \ 
{V}^{K-1}_i(x,y,b,\omega_i) \le {V}^K_i(x,y,b,\omega_i).$$
Moreover, $\ubar{V}^K_i(x, y, b, \omega_i)$ is convex in $(x,y)$ for fixed $b$ and concave in $b$ for fixed $(x,y)$. And,
$$
\mathbb{E}_{\omega_i^m} \ubar{V}^K_i(x,y,b,\omega_i) \ge \max_{k \in \cK} \left [ \alpha_{i,m,k} + \beta_{i,m,k}^\top \, x + \gamma_{i,m,k}^\top \, y \right ] ,
$$
where $\alpha_{i,m,k}, \beta_{i,m,k}, \gamma_{i,m,k}$ are computed by~\eqref{eqn:cut_coefficients_ddu_y_learn} for $i \in \mathcal{N}$, $m \in \mathcal{M}$, $k \in \cK$. 
\end{theorem}

Algorithm~\ref{alg:ddu_sddp_dd_learning} gives the $K$-th iteration of our variant of SDDP.  We solve non-convex subproblems $\mathbf{SP}^{K-1}_i$ in forward passes and linear programs $\underline{\mathbf{SP}}^{K-1}_i$ in backward passes. The forward pass accounts for our belief state, and its updates, when sampling and when solving $\mathbf{SP}^{K-1}_i$. In the backward pass subproblems $\underline{\mathbf{SP}}^{K-1}_j$ are identical for each model $m$, and only when computing cuts must we account for $\mathbb{P}_j^m$ via equation~\eqref{eqn:cut_coefficients_ddu_y_learn}. We include~$b^K$ when adding cuts to $\mathbf{SP}^K_i$ and $\underline{\mathbf{SP}}^K_i$ because of the $b_{m,k}$ terms in the resulting inequalities. 

\begin{algorithm}[!ht]
\scriptsize
    \SetAlgoLined
    \tcc{Forward pass}
    set $\bar{x}=x_R$, $b=b_R$, $\mathcal{S} = [\;]$, $i = R$, $\bar{y} = 0$, and $\bar{y}^\prime \in \mathcal{Y}$ \\
    \While{$rand() < \sum{}_{m \in \mathcal{M}, j \in i^+} \, b_m \phi^m_{ij}(\bar{y}) 
    $}{
        set $i^-=i$ \\
        sample new $i$ from $i^+$ according to pmf, $\frac{\sum{}_{m \in \mathcal{M}} \, b_m \phi^m_{i, \cdot}(\bar{y})}{\sum_{m \in \mathcal{M}, {j} \in i^+} b_m \phi^m_{ij}(\bar{y})}$ \\
        sample $\omega$ from $\Omega_i$ according to pmf mixture $\sum_{m \in \mathcal{M}} b_m \mathbb{P}^m_i$ \\
        set $b'=B(\bar{y},b,i^- \rightarrow i,\omega)$ using~\eqref{eq:belief_update_dec_dep} \\
        solve \textbf{SP}${}_i^{K-1}(\bar{x},\bar{y}^\prime,b',\omega)$ and obtain $x^\prime$ and $y^\prime=[y_j^\prime]_{j \in i^+}$ \\
        append $(i, x^\prime, y^\prime, b^\prime)$ to the list $\mathcal{S}$\\
        set $\bar{x} = x^\prime$, $\bar{y} = \bar{y}^\prime$, $\bar{y}^\prime = y^\prime$, $b=b^\prime$\\
    }
    \tcc{Backward pass}
    \For{$(i,\bar{x}^K,\bar{y}^K,b^K)$ in $reverse(\mathcal{S})$}{
        \If {$i^+ = \varnothing$}{
            continue to next {\bf for}-loop iterate of $reverse(\mathcal{S})$ \\
        }
        \For{$j\in i^+$}{
        \For{$\omega \in\Omega_j$}{
            solve $\underline{\mathbf{SP}}_j^{K-1}(\bar{x}^K,\bar{y}^K_j,b^K,\omega)$ \\
            set $\ubar{V}^{K-1}_j(\bar{x}^K,\bar{y}^K_j,b^K,\omega)$ to the optimal objective function value\\
            {{set $\bar{\lambda}^K_{x,j,\omega}$ and $\bar{\lambda}^K_{y,j,\omega}$ to optimal dual variables $\lambda_x$ and $\lambda_y$}} \\
        }
        \For{$m \in \mathcal{M}$}{
                {{ 
        compute cut coefficients $\alpha_{j,m,K}, \beta_{j,m,K},\gamma_{j,m,K}$ using equation~\eqref{eqn:cut_coefficients_ddu_y_learn}}} \\
        {add $K^{th}$ cut to \textbf{SP}${}_{i}^{K}$
        and $\underline{\mathbf{SP}}_{i}^{K}$
        using $\alpha_{j,m,K}, \beta_{j,m,K},\gamma_{j,m,K}$ and $b^K$}
        }
        }
    }
    \caption{$K^{th}$ iteration of SDDP under recursion~\eqref{eq:belief_sp_learn}.}\label{alg:ddu_sddp_dd_learning}
\end{algorithm}

\section{Computational examples}\label{sec:computational-results}

As a computational demonstration of our algorithms, we solve two examples in \texttt{SDDP.jl}, a free and open-source Julia \cite{bezanson_julia_2017} package for modeling policy graphs and solving them with stochastic dual dynamic programming \cite{dowsonSDDPJlJulia2017}. \texttt{SDDP.jl} is based on the JuMP modeling language \cite{Lubin2023}. The code and data for our experiments are available at \url{https://github.com/odow/SDDP.jl}. Our intent is neither to demonstrate large-scale examples nor to implement the adaptive decision-dependent learning algorithm; we leave that to future work.

\subsection{The cheese producer}

\begin{figure}[!ht]
    \centering
    \includegraphics[width=0.70\textwidth]{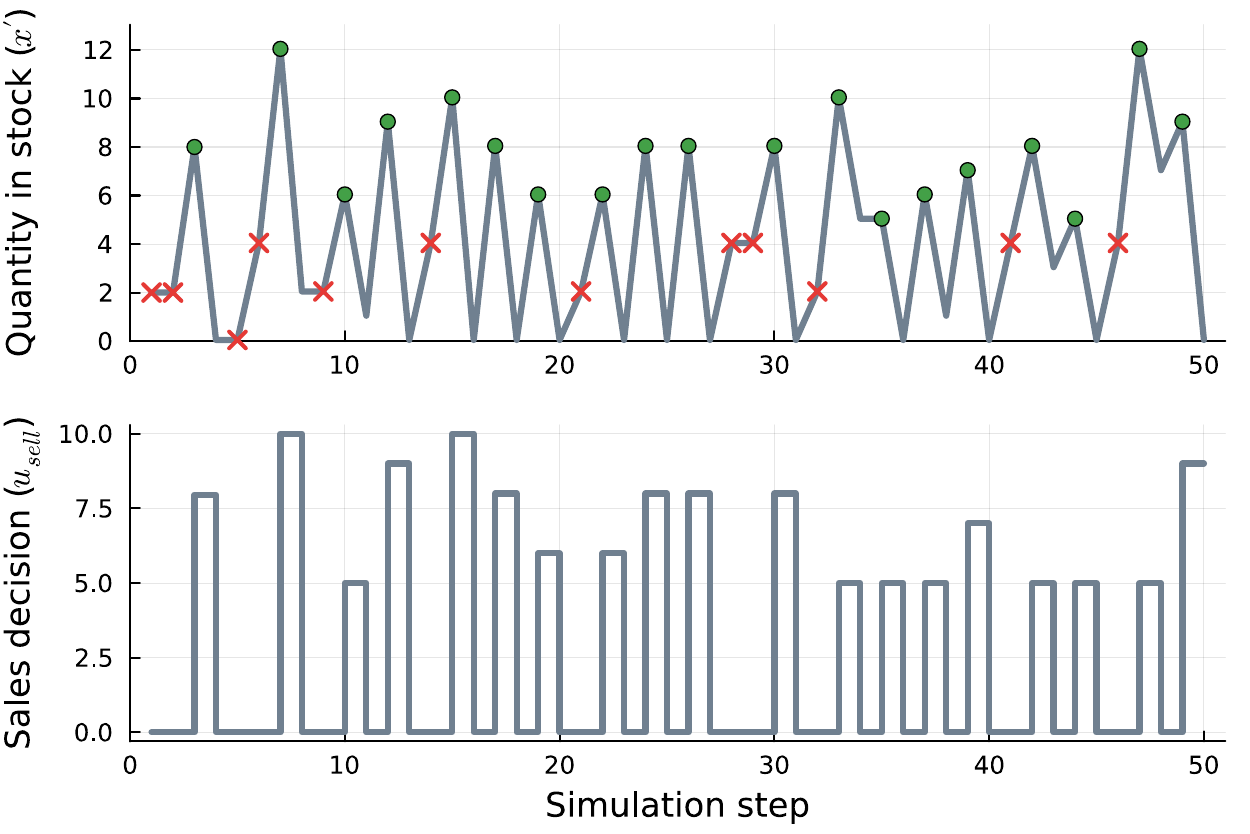}
    \caption{One simulation trajectory of the cheese producer policy over 50 time-steps. The red crosses correspond to states when $y = 0$ and the farmer does not visit the market, and the green circles correspond to states when $y = 1$ and the farmer visits the market.}
    \label{fig:computation-cheese-producer}
\end{figure}

To demonstrate the Lagrangian relaxation algorithm from Section~\ref{sec:lagrangian_relaxation}, we solve Example~\ref{example:cheese}. As data, we use $\rho = 0.9$, $c_y = 3$, $c_u = 1$, $\Omega_q = \{0, 2, 4, 6, 8\}$, $\Omega_d = \{5, 10\}$, each with a uniform pmf. One sample path of the policy is shown in Figure~\ref{fig:computation-cheese-producer}. The policy appears to correspond to the decision rule that $y = 1$ if $x^\prime \ge 5$ and $y = 0$ if $x^\prime < 5$.

\begin{figure}[!ht]
    \centering
    \includegraphics[width=0.70\textwidth]{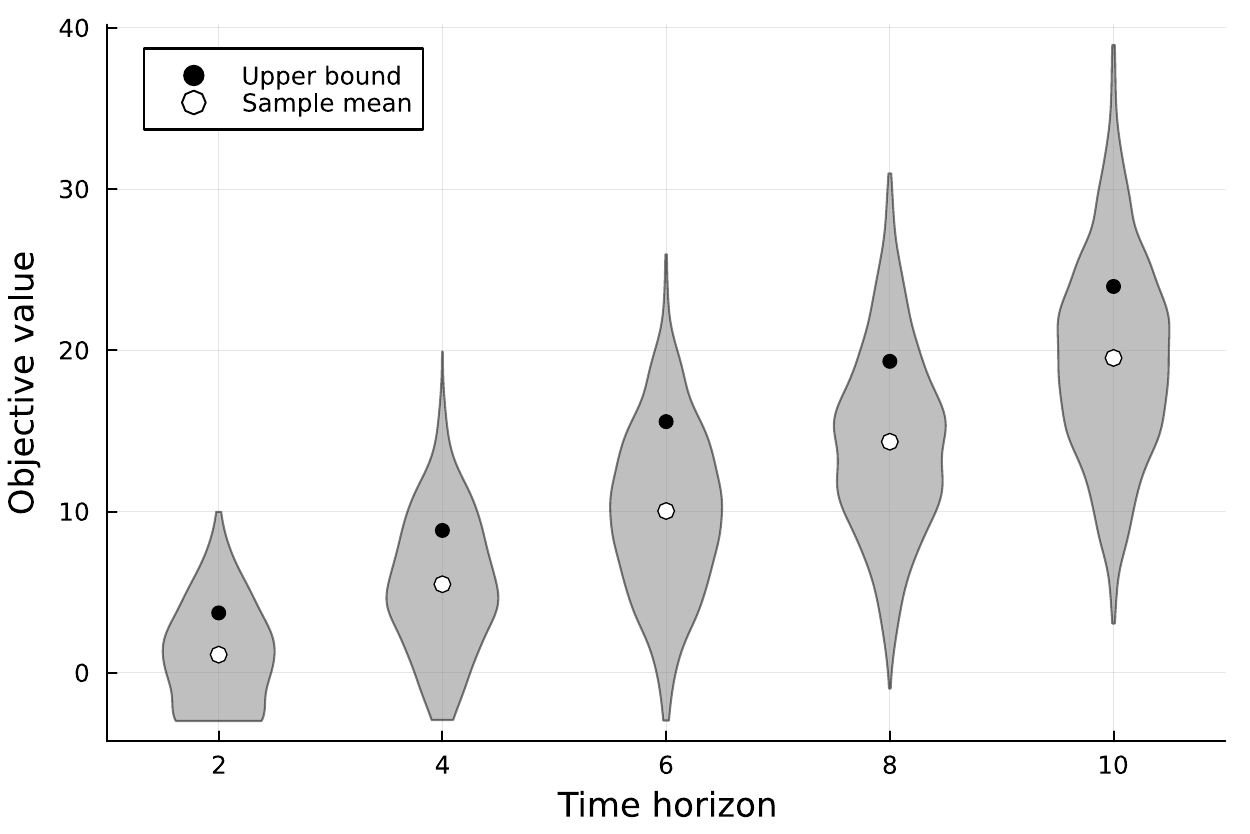}
    \caption{Violin plots of the distribution of 1000 simulated objective values for Example~\ref{example:cheese} with differing finite time horizon $T$. The black dot corresponds to the upper bound on expected profit obtained from SDDP. The white dot is the sample mean of the distribution.}
    \label{fig:computation-cheese-producer-violin}
\end{figure}

We further analyze Example~\ref{example:cheese}, reformulating it with a finite-horizon of $T=$ 2 to 10 weeks. Each week consists of the farm node and the optional market node. We train each policy for 200 iterations of SDDP and then perform 1000 simulations of the policy.  Figure~\ref{fig:computation-cheese-producer-violin}'s violin plots show the distribution of simulated objective values, where we maximize profit. Due to the non-convex value function, our upper bound does not converge to the sample mean. However, if we have small gaps, it gives us confidence that the policies are good.

\subsection{The tiger problem}

\begin{figure}[!ht]
    \centering
    \includegraphics[width=0.70\textwidth]{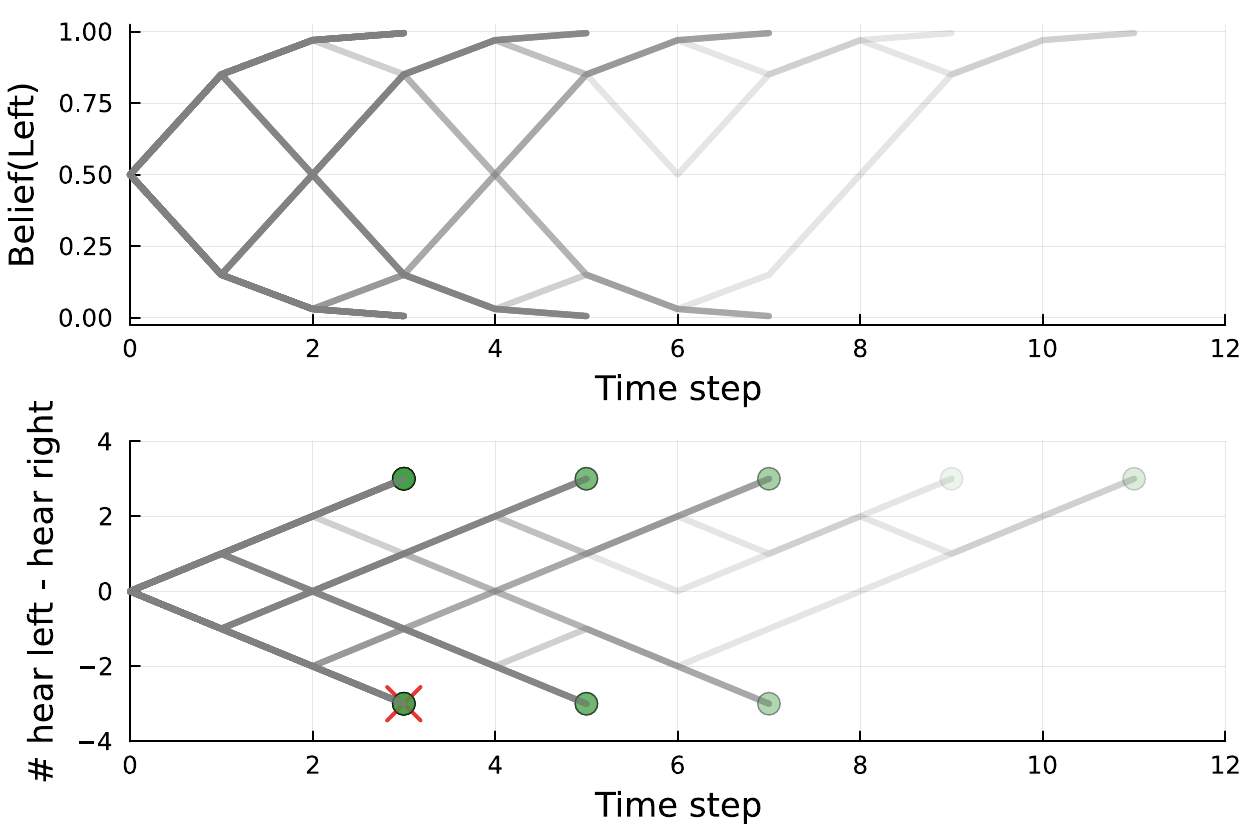}
    \caption{One hundred simulations of the tiger policy with a false positive rate of 15\%. The green circles correspond to ``correctly'' opening the door to escape. There is one red cross at position $(3, -3)$, which is a simulation that ``wrongly'' opened the door to the tiger after hearing three consecutive false positives.}
    \label{fig:computation-tiger}
\end{figure}

To demonstrate our decision-dependent statistical learning model, we solve Example~\ref{example:tiger} using Algorithm~\ref{alg:ddu_sddp_dd_learning} but with the Lagrangian relaxation from Section~\ref{sec:lagrangian_relaxation}. We use a discount factor of $\rho = 0.95$, and we use a false positive rate of 15\%. Figure~\ref{fig:computation-tiger} shows 100 simulations of the belief that the tiger is behind the left door over the number of time steps, as well as the net count of the number of times we hear the tiger behind the left door. The policy corresponds to the decision rule of opening the opposite door once the net count reaches three. Of the 100 simulations, there is one example in which the agent opens the ``wrong'' door to the tiger after hearing three consecutive false positives.

\begin{figure}[!ht]
    \centering
    \includegraphics[width=0.70\textwidth]{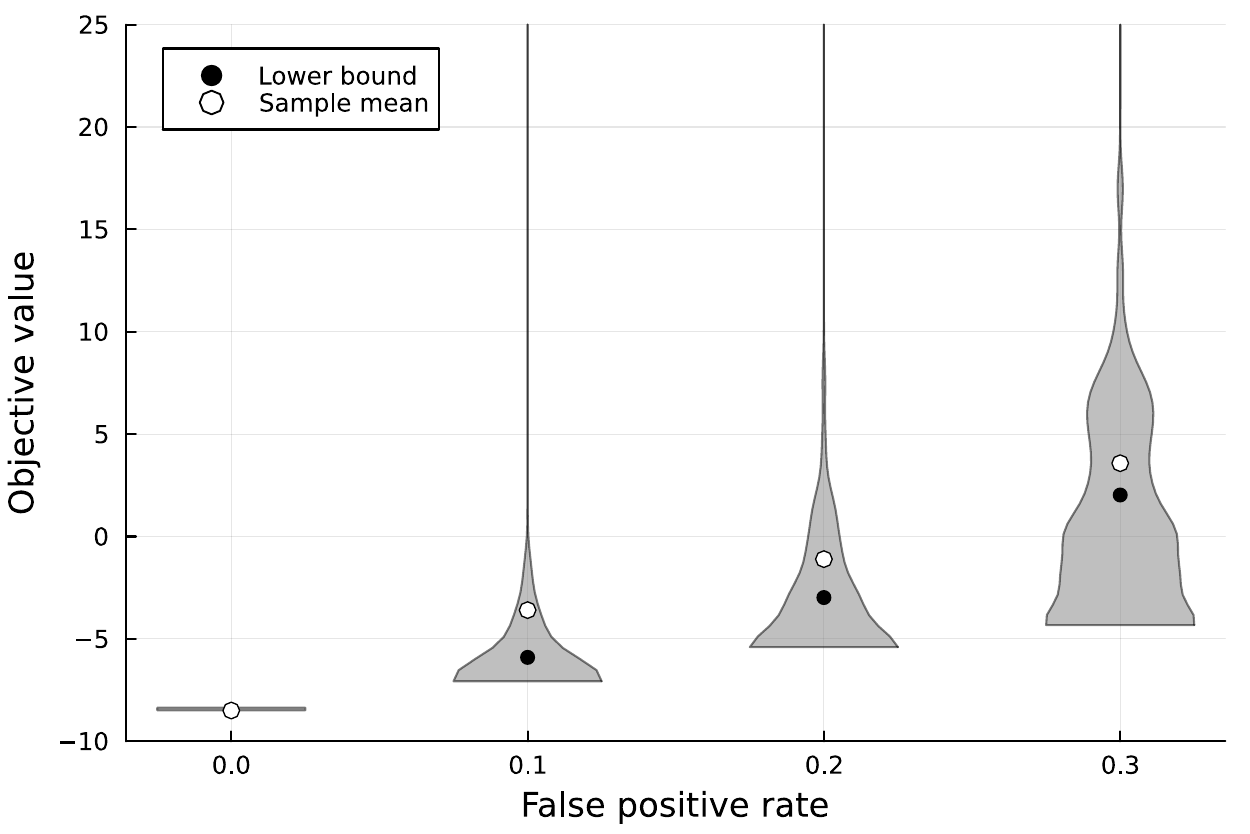}
    \caption{Violin plots of the distribution of 100 simulated objective values for the tiger problem with varying false positive rates. The black dot corresponds to the lower bound of the expected value obtained from SDDP. The white dot is the sample mean of the distribution. The thin upper tails (full extent not shown) represent simulations in which the door to the tiger was opened and a cost of \$100 was incurred.}
    \label{fig:computation-tiger-violin}
\end{figure}

We further analyze Example~\ref{example:tiger} via a parametric study for false positive rates of 0\% (perfect hearing), 10\%, 20\%, and 30\%. We train each policy for 100 iterations and then perform 100 simulations of the policy. Figure~\ref{fig:computation-tiger-violin} shows violin plots of the simulated objective values. As expected, as the false positive rate grows, the agent has a harder time learning the location of the tiger, and the cost increases.  Because of the non-convex cost-to-go function, our lower bound does not converge to the sample mean. However, if we have small gaps, then we have confidence that the policies are good in practice.

\section{Conclusion}

\revision{We have shown how policy graphs} can model a range of sequential decision problems under uncertainty. From the viewpoint of stochastic programming, we have extended policy graphs to incorporate ideas from MDPs, including decision-dependent transition probabilities, a limited form of statistical learning, and their combination: decision-dependent learning. From the viewpoint of the MDP literature, we have presented policy graphs and SDDP algorithms as a practical combination for modeling and solving a class of structured MDPs with continuous states and continuous actions, \revision{which we call CACS MDPs.}

Our model for decision-dependent learning allows the agent {to take actions that} reveal information so that better decisions can be subsequently made. Our models for both decision-dependent transitions and decision-dependent learning yield non-convex cost-to-go functions. Our algorithmic variants of SDDP use a mix of convex relaxations to form a policy, and the original non-convex models to evaluate the policy's performance.  

Although our focus has been on modeling and algorithmic development, we provide an open-source implementation of our ideas in \texttt{SDDP.jl} \cite{dowsonSDDPJlJulia2017}. The code and documentation are freely available at \url{https://sddp.dev}. As future work, we will conduct larger-scale computational experiments and integrate our work with other features supported by \texttt{SDDP.jl}, such as risk aversion.

\section*{Acknowledgments}
\revision{The authors thank two anonymous referees for comments that improved the paper.} David Morton's research was supported, in part, by the US Department of Homeland Security (DHS) under Grant Award Number \mbox{17STQAC00001-06-00}. The views and conclusions contained in this document are those of the authors and should not be interpreted as necessarily representing the official policies, either expressed or implied, of DHS.

\end{document}